%% file: iclr2027_conference.tex
\documentclass{article} 
\usepackage{iclr2027_conference,times}

\input{math_commands.tex}

\usepackage{hyperref}
\usepackage{url}
\usepackage{graphicx}
\usepackage{booktabs}
\usepackage{amsmath, amssymb}
\usepackage{hyperref}
\usepackage{multirow}

\newcommand{\ds}[1]{\texttt{#1}}

\title{Dataset identity, not novelty: the source of \\ an inflated OOD detection gain}

\author{Donghoon Lee \\
Hongik University\\
\texttt{ddozmi@mail.hongik.ac.kr}
\And
Shinjin Kang\thanks{Corresponding author.}\\
Hongik University\\
\texttt{directx@hongik.ac.kr}
}

\iclrfinalcopy 
\begin{document}

\maketitle

\begin{abstract}
A post-hoc out-of-distribution (OOD) detector reads the activations of a trained classifier and returns a score. It fits that score on in-distribution data, and the benchmarks that evaluate it supply a second piece of OOD data for the fitting itself. Some detectors tune a constant on it. Others fit a direction in feature space or train a flexible combiner and report the number that it reaches as the gain that is still available. Every one of these fits goes through the same check, which validates it on held-out samples of the same OOD dataset. That check rules out memorizing individual images. It says nothing about a fit that has instead learned which dataset it is looking at, and a direction that recognizes one OOD dataset rather than novelty passes it perfectly. The detector that a practitioner installs meets OOD data from a source that nobody fitted it on, so the difference decides what the reported number is worth. We measure it by holding out the whole OOD dataset rather than a sample of it, and we call that gap the inflation. We read it across a range of combiners on ImageNet backbones and CIFAR-100 architectures. Most of the gain that the usual protocol reports turns out to be dataset identity rather than novelty. The size of the fit does not move what survives, so the effect is not ordinary overfitting. The share depends instead on whether the input exposes class identity, and two controls that vary that property alone separate the inflation on every backbone of both benchmarks. A closed form accounts for the effect and computes it from the fitting rows, so a practitioner can tell which fits will inflate without running the hold-out protocol at all. One of these fits survives, namely the single constant that the field already picks on a designated validation dataset. Anything above it reports a gain that the hold-out protocol does not return, and on one benchmark what survives falls while what is reported climbs.
\end{abstract}

\section{Introduction}
\label{sec:introduction}

\begin{figure}[htbp]
\centering
\includegraphics[width=\textwidth]{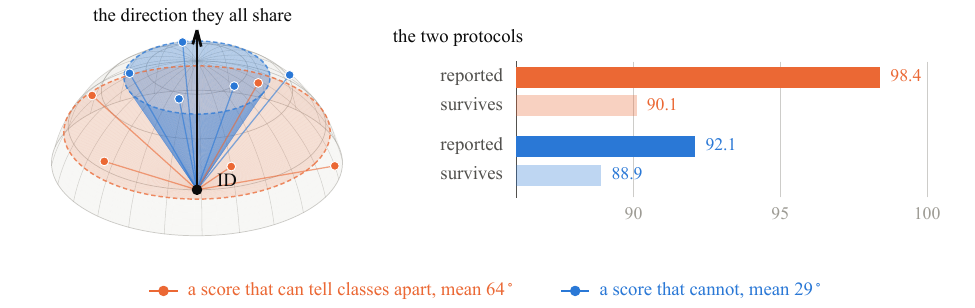}
\caption{Where the OOD datasets point. The usual protocol gives the fit the direction of the dataset that it will be scored on, while deployment gives it the direction that the other datasets share. The pole is that shared direction and the dome is the unit sphere around it. Each dot is one OOD dataset at its angle to the pole, for the two profiles at full dimension over 11 ImageNet backbones. Each cone opens to its family mean. The bars give the AUROC each family reaches under the two protocols. Rotation around the pole carries no data.}
\label{fig:concept}
\end{figure}

A post-hoc out-of-distribution (OOD) detector reads the activations of a trained classifier and returns a score. It fits that score on in-distribution data and then reports it against several OOD datasets. Benchmarks also supply OOD data for the fitting itself. OpenOOD \citep{zhang2024openood} sets aside a validation split, and papers that reproduce a family of detectors under one protocol pick their hyperparameters on that split \citep{kim2026mahavar}. Some pick a temperature and a perturbation size on OOD samples \citep{liang2018odin}. Some regress one weight per layer on in-distribution and OOD images, so they carry a set of weights per OOD dataset \citep{lee2018mahalanobis}. A separate line fits one combiner over several such scores and reports the gain that follows \citep{novello2024combining,magesh2023multiple}. Every one of these fits goes through the same check. The check validates the fit on held-out samples of the same OOD dataset.

That check removes one failure mode and misses a second one. Held out samples rule out memorizing individual images. They say nothing about a fit that has learned which dataset it is looking at. A direction that recognizes one OOD dataset rather than novelty will still transfer perfectly between two halves of that dataset. Sample level validation therefore certifies sample level generalization and nothing beyond it. What sets the size of that second failure mode has not been measured. Domain generalization already named the problem. \citet{gulrajani2021domainbed} give the name oracle selection to any choice that reads the test domain, and they ask for results obtained that way to be disclaimed. The rule has not been carried into OOD detection, though the symptom it guards against is on record there. Figure~\ref{fig:concept} states the question geometrically and gives our answer to it.

We measure it by holding out the OOD dataset itself. We fit the detector on some OOD datasets and then score it on a dataset that it has never seen. The difference between the two protocols is the part of the reported gain that comes from dataset identity. We apply this to the sequence of combiners that Table~\ref{tab:capacity} lays out.

Dataset identity accounts for most of what the usual protocol reports. The gap between the two protocols widens as the fit gains directions, and CIFAR-100 loses more at every step.

The size of the fit is not what moves the column that survives. Appendix~\ref{app:mlpwidth} moves the parameter count over four orders of magnitude at a fixed input. The reported column rises over that range and the held-out column does not. Width does raise the inflation, and Section~\ref{sec:exp-mech} identifies the factor responsible. Our channels are permutation invariant functions of the class distances, so none of them can report which class a sample is near. Random directions of the feature space carry the same width and the same parameter count, and those directions do report it. A fit on them inflates several times as much on every backbone of both benchmarks.

A closed form says why. A discriminant points along the displacement of the OOD set that the fit was given, and the gap between the two protocols is the length of that displacement times an angular factor. Width sets the length and what the input exposes sets the angle. Both factors read the fitting rows and neither needs a hold-out pass. Their product ranks the measured inflation almost exactly. At matched width the invariant channel inventory has the longer displacement and still inflates less, because the invariant cosine is higher.

One step is exempt. The field picks a single constant on the OOD dataset that the benchmark designates for validation, and we call that condition validation tuned. Neither benchmark shows any inflation there, and no backbone moves far in either direction. Choosing one number this way is sound at that capacity. Every step above it fails.

Measuring any of this needs a zero point that nobody tuned. We use a score with no free constant at all and call it VarRatio. The score needs no OOD data and no published constant, and any rescaling of the whitened space leaves the ranking untouched. We introduce VarRatio as an instrument, and Section~\ref{sec:exp-ref} shows only that the score is strong enough to serve as one. VarRatio improves on Mahalanobis++ \citep{muller2025mahalanobispp} on 16 of our 17 backbones. Without any fitting the score comes within $0.22$ of a ridge discriminant that was honestly fitted on four OOD datasets.

The reported number and the number that survives move apart as the sequence widens. On ImageNet what survives climbs by far less than what is reported. On CIFAR-100 what survives falls while what is reported climbs.

\paragraph{Contributions.}
\begin{itemize}
\item A measurement that splits a reported gain in two. One part survives a change of OOD dataset and the other does not. We read both against a zero point that has no constant to publish. Running the measurement across combiners shows that the parameter count moves the split far less than the material does.
\item Two controls that vary class identity alone. Sorted and unsorted distance profiles agree in dimension, content and fitting procedure. Random directions of the feature space match the channel inventory in width and in parameter count.
\item A closed form for the effect. The inflation is a displacement length times an angular factor. The width of the fit sets the first factor and what the input exposes sets the second. Both are computable from the fitting rows.
\end{itemize}

\section{Related work}

Post-hoc detectors differ in what they read before they can score. One family reads the classifier head alone \citep{hendrycks2017msp,hendrycks2022maxlogit,liu2020energy,liu2023gen}. A second reads the penultimate features with the class labels \citep{lee2018mahalanobis,ren2021relative,muller2025mahalanobispp,kim2026mahavar}. Others read the features and the head together \citep{wang2022vim,demirel2025ora}, keep a bank of training features \citep{sun2022knn}, or carry a percentile that reshapes the activations \citep{sun2021react,djurisic2023ash}. A further line fits one combiner over several such scores \citep{novello2024combining,magesh2023multiple}. We sort the comparison by what a detector needs before it can score, because a feature bank or a tuned constant is out of reach for a practitioner who has neither.

Benchmarks decide how those constants get their values. OpenOOD fixes the splits and the near and far grouping for both benchmarks that we use, and it supplies a separate OOD validation split \citep{zhang2024openood}. Papers pick every hyperparameter on that split \citep{kim2026mahavar}, and the practice is older than the benchmark \citep{liang2018odin,lee2018mahalanobis}. Later work removes the dependence \citep{hsu2020godin,mondal2026tuning,techapanurak2020hyperparameterfree}. The rule that Section~\ref{sec:introduction} borrows comes from domain generalization \citep{gulrajani2021domainbed,zhang2023nicopp}. The symptom is already on record in OOD detection, in benchmarks where no detector leads across dataset pairs \citep{tajwar2021notrue}, in protocol work that scores a detector on a source the fit never saw \citep{shafaei2019odtest,hendrycks2019oe}, and in the fading advantage of outlier exposure as a benchmark grows \citep{wang2025dissecting}. Theory gives no general guarantee that a rule fitted against one OOD distribution carries to another \citep{fang2022learnable}.

\section{Preliminaries}
\label{sec:prelim}

\paragraph{Notation.}
A classifier maps an image to a penultimate feature $\phi(x) \in \mathbb{R}^{D}$ and then to $C$ logits. Following Mahalanobis++ \citep{muller2025mahalanobispp} we normalize the feature to unit length and write $\hat{\phi}(x) = \phi(x) / \lVert \phi(x) \rVert_2$. The training split gives a class mean $\mu_c$ for every class and one tied covariance $\Sigma$. We whiten once and write $z = \Sigma^{-1/2}\hat{\phi}$ and $\mu_c^{w} = \Sigma^{-1/2}\mu_c$, which turns the squared Mahalanobis distance into a squared Euclidean distance
\begin{equation}
d_c(x) \;=\; \lVert z(x) - \mu_c^{w} \rVert_2^{2}, \qquad c = 1, \dots, C .
\label{eq:dist}
\end{equation}
We drop the superscript from now on. The vector $d(x) = (d_1, \dots, d_C)$ is the distance profile. Two readouts of it appear throughout. The first is the nearest class distance $d_{(1)} = \min_c d_c$, which is the Mahalanobis++ score. The second is the variance of the profile $\operatorname{Var}_c d_c$, which MahaVar adds to it \citep{kim2026mahavar}.

\paragraph{Permutation invariance.}
Class labels index the profile but carry no order. A readout $f$ of the profile is permutation invariant when
\begin{equation*}
f(d_1, \dots, d_C) \;=\; f\big(d_{\pi(1)}, \dots, d_{\pi(C)}\big)
\qquad \text{for every permutation } \pi \text{ of the } C \text{ classes},
\end{equation*}
so that $f$ sees the multiset $\{d_1, \dots, d_C\}$ and never which class produced which entry. Both $d_{(1)}$ and $\operatorname{Var}_c d_c$ have that property. The sorted profile $\operatorname{sort}(d)$ has it too and stands as the canonical invariant. The raw vector $d$ is not invariant. We call whatever a combiner reads its material. A fit that reads $d$ can respond to the distance of one named class and a fit that reads $\operatorname{sort}(d)$ cannot, even though the two hold the same $C$ numbers.

\paragraph{Units.}
Equation~\ref{eq:dist} squares a length, so $d_c$ carries units of $L^{2}$ and $\operatorname{Var}_c d_c$ carries units of $L^{4}$. A score of the form $d_{(1)} - \alpha \operatorname{Var}_c d_c$ therefore needs $\alpha$ to carry units of $L^{-2}$. The whitened length is what the covariance estimate sets, and rescaling it exposes the constant directly. Sending $z \to \lambda z$ gives
\begin{equation}
d_{(1)} - \alpha \operatorname{Var}_c d_c \;\longmapsto\; \lambda^{2}\Big( d_{(1)} - \alpha \lambda^{2} \operatorname{Var}_c d_c \Big),
\label{eq:rescale}
\end{equation}
so the ranking depends on the product $\alpha\lambda^{2}$ and never on $\alpha$ alone. Dividing the constant by the scale that it belongs to removes the dependence. A paper that reports the constant in dimensionless form therefore reports a number that another group can use. Section~\ref{sec:score} returns to this and Appendix~\ref{app:units} gives the form and the measurements.

\section{Measuring what a fit buys}

\subsection{Two protocols}
\label{sec:protocol}

Write $T$ for the set of OOD datasets in a benchmark. We split the ID validation rows into halves $A$ and $B$, and we split each OOD dataset $t$ into halves $t_a$ and $t_b$. A fitting rule takes ID rows and OOD rows and returns a score, and we write $\hat{s}_{X}$ for the score that it returns from the rows $X$. Two protocols use the same rule and differ only in which OOD rows they hand it,
\begin{align}
w(t) &\;=\; \operatorname{AUROC}\Big( \hat{s}_{A \cup t_a} \;\Big|\; B, \, t_b \Big),
\label{eq:within}\\[2pt]
h(t) &\;=\; \operatorname{AUROC}\Big( \hat{s}_{A \cup \bigcup_{u \neq t} u_a} \;\Big|\; B, \, t_b \Big).
\label{eq:holdout}
\end{align}
We call $w$ the within protocol and $h$ the hold-out protocol. The inflation is their gap averaged over the benchmark,
\begin{equation*}
I \;=\; \frac{1}{|T|} \sum_{t \in T} \big( w(t) - h(t) \big).
\end{equation*}

The two protocols share their evaluation rows. They also share the ID rows that they fit on. The only thing that changes is which OOD dataset the fit has seen, so the difference between them measures what the fit gained from seeing that dataset rather than from seeing OOD data at all. We call that difference the inflation.

\paragraph{Two ways to tune one constant.}
Equation~\ref{eq:within} lets the fit see the fit half of the dataset that it will be scored on. That is the sharpest form of the within protocol and we call it target tuned. The field runs a milder form. OpenOOD \citep{zhang2024openood} designates one OOD dataset as a validation set, and a paper picks its constant on that dataset and then reports on every test set \citep{kim2026mahavar}. We call that condition validation tuned. The two differ in how much of the target dataset reaches the fit, so we measure them separately and report both in Section~\ref{sec:exp-capacity}. Appendix~\ref{app:protocol} gives the grids and one caveat that the designated validation set carries.

The hold-out protocol is the deployment condition at the fitting suite that the benchmark supplies. A detector that a practitioner installs meets OOD data from a source that nobody fitted it on. Cross-validation that leaves out one dataset at a time is a known idea and we do not claim it \citep{zhang2024openood}. What we add is a reading of the inflation against how much freedom the fit has.

\subsection{The zero point}
\label{sec:score}

Reading the inflation needs a zero point that nobody tuned. A tuned one makes the comparison circular and a weak one makes every fitted gain look larger than it is. We need it to carry no constant and to be strong enough that a gain measured against it is a real gain. We use
\begin{equation}
S(x) \;=\; \log \min_c d_c(x) \;-\; \log \operatorname{Var}_c\!\big(d_c(x)\big),
\label{eq:varratio}
\end{equation}
which we call VarRatio. It reads the two quantities that the Mahalanobis family already computes and takes their log ratio. It needs the training features and the class labels and nothing else. Appendix~\ref{app:identity} says what the denominator reads and why the class mean subspace sets it. There is no OOD data in it, no validation split and no number that anyone has to publish.

Rescaling the whitened space leaves the ranking untouched. Sending $z \to \lambda z$ sends $d_c \to \lambda^{2} d_c$ and $\operatorname{Var}_c d_c \to \lambda^{4} \operatorname{Var}_c d_c$, so
\begin{equation*}
S \;\longmapsto\; S - 2 \log \lambda
\end{equation*}
for every sample at once. AUROC does not see an additive constant, which is the protection that \eqref{eq:rescale} denies the linear form. The mixing constant of the linear form has no such protection, and Appendix~\ref{app:units} measures what that costs.

The exponent on the second term is a choice and we set it to 1. Invariance does not force that value, because any exponent shifts the score by a constant. We fix it because tuning it would require publishing it, and Appendix~\ref{app:exponent} shows that sweeping it against the test sets buys almost nothing.

\section{Experiments}
\label{sec:exp}

\subsection{Setup}

ImageNet-1k \citep{russakovsky2015imagenet} runs on 11 public checkpoints and we never train one. CIFAR-100 \citep{krizhevsky2009cifar} has no public checkpoint of this kind, so we train 6 architectures ourselves under one frozen recipe and repeat each one under 3 seeds. Appendix~\ref{app:repro} gives the recipe, the accuracy of every run and the three places where our CIFAR data differs from the OpenOOD release.

We implement every baseline ourselves and put all of them through one protocol. Appendix~\ref{app:repro} gives the four baselines whose constant we set by an oracle on the very test sets that we then score, which is a handicap in their favour.

\paragraph{The sequence of combiners.}
\label{sec:sequence}
We index a combiner by the number of directions that it may look at, and Table~\ref{tab:capacity} names every step and gives the parameters that it holds. The smallest step tunes one constant $\alpha$ in $d_{(1)} - \alpha \operatorname{Var}_c d_c$ over a fixed grid. The middle steps fit a ridge discriminant on prefixes of a fixed channel inventory. The last steps replace that input by the whitened feature $z$ and replace the ridge by a network with two hidden layers, so the table crosses width against parameter count. The inventory holds 17 channels and we keep the 13 that both benchmarks define in the same way, which puts the two benchmarks on one axis. We take the channel order from an inventory we built long before we asked this question, so the prefixes nest and we never chose them to suit the result. The ridge strength comes from a nested split inside the fitting half and never from the evaluation rows. Appendix~\ref{app:channels} lists the channels.

\paragraph{Two controls that vary class identity alone.}
\label{sec:control}
The sequence changes width, parameter count and representation together, so two controls hold everything but one property fixed. The first holds the content fixed as well. The sorted profile $[\operatorname{sort}(d), \lVert z \rVert^{2}]$ and the unsorted profile $[d, \lVert z \rVert^{2}]$ carry the same $C + 1$ numbers in the same dimension and take the same ridge discriminant, and only the sorted one hides which class produced which distance. The second fits the same ridge discriminant on $m$ random orthonormal views of the whitened feature $z$, for $m$ from 2 up to $D$. At $m = D$ the projection is a rotation, so the top of the sweep reproduces a ridge on $z$ itself and the sweep has a verified anchor. Read at $m = 13$ it carries exactly the width and the parameter count of the 13 channel step, and it differs only in that a linear view of $z$ can respond to the distance of a named class while a permutation invariant channel cannot. We read both controls through the same random projections, so they meet at every width. Appendix~\ref{app:proj} gives both constructions.

\subsection{The reference}
\label{sec:exp-ref}

Every inflation in this paper is a gap read against VarRatio, and Appendix~\ref{app:ref} carries three checks that say the reference is not weak. It improves on Mahalanobis++, the score that it modifies, on 16 of our 17 backbones. Among detectors that use no OOD data it holds the highest worst case backbone on both benchmarks. A ridge discriminant on the same two quantities, fit on four OOD datasets and scored on the fifth, beats it by $0.22$ on ImageNet.

\begin{figure}[t]
\centering
\includegraphics[width=\textwidth]{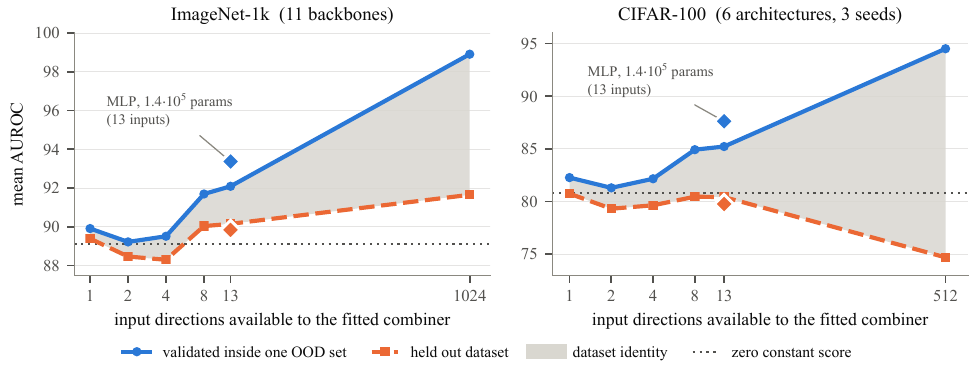}
\caption{What a fitted score buys and what survives. The solid line gives the AUROC that the usual protocol reports and the dashed line gives the AUROC that survives once we hold the whole dataset out. The distance between them is the part of the reported gain that comes from dataset identity. The dotted line marks the zero capacity reference, which fits nothing. Both panels run on the 13 shared channels, and the widest point adds the whitened feature space.}
\label{fig:capacity}
\end{figure}

\subsection{What a fit buys and what survives}
\label{sec:exp-capacity}

Figure~\ref{fig:capacity} and Table~\ref{tab:capacity} carry the main result. The two protocols agree at the bottom of the sequence and separate steadily as the fit gains directions. CIFAR-100 pays more than ImageNet at every step of the channel sequence. At the top of the ImageNet sequence three quarters of what the sequence adds to the reported column is missing from the held-out column.

Neither axis that the sequence offers explains that curve. Appendix~\ref{app:mlpwidth} sweeps the width of a net on the same 13 channels, so the parameter count runs from 13 to $5.4 \cdot 10^{5}$. The inflation rises by $1.61$ over that range and every step adds less than the one before. The hold-out arm moves by $-0.34$, so what the parameters buy stays inside the dataset. The widest linear fit inflates more than the network that reads 13 channels, while holding a small fraction of its parameters.

Width behaves differently, and reading it takes one piece of care. In AUROC the inflation peaks well before full width and then falls, which is a property of the scale rather than of the fit. We report the AUROC curve because that is the number that a paper would print, and Section~\ref{sec:mech} reads the same sweep where it has no ceiling.

One step escapes the pattern and it is the one the field actually uses. The validation tuned row of Table~\ref{tab:capacity} picks a single constant on the validation dataset that OpenOOD designates, and it inflates by nothing on either benchmark. Target tuning the same constant on the dataset that it will be scored on costs the first row of the sequence instead, so what the constant can absorb depends entirely on where it was picked. Selecting one number on a designated validation dataset is sound practice at that capacity. What fails is every step above it.

How much OOD data the hold-out arm may read changes the inflation without removing it. Appendix~\ref{app:ksweep} grows the fitting suite one dataset at a time, and the inflation falls throughout and is still falling at the largest suite either benchmark offers. The two columns of Table~\ref{tab:capacity} move apart as the sequence widens. From the zero capacity row to the linear map on the feature space the reported column climbs by $9.78$ on ImageNet and the held-out column climbs by $2.52$. On CIFAR-100 the reported column climbs by $13.67$ and the held-out column falls by $6.14$. Per backbone the split is sharper than the mean suggests, and Figure~\ref{fig:loofail} in the appendix resolves it. Only the linear map on the feature space lifts every ImageNet backbone above the row the sequence starts from, and no step lifts every CIFAR-100 architecture.

\begin{table}[t]
\centering
\caption{What each fit reads, how many parameters it holds, and what survives. Columns give the AUROC that the usual protocol reports, the AUROC that survives holding the OOD dataset out, and the difference. The parameter column counts the ImageNet fit at $D = 1024$, and counts that depend on $D$ vary with it. Every row scores the held-out halves of Section~\ref{sec:protocol}, so the reference row differs from Table~\ref{tab:main}, and Appendix~\ref{app:spread} gives the spread behind each average.}
\label{tab:capacity}
\small
\setlength{\tabcolsep}{4pt}
\begin{tabular}{lrcccccc}
\toprule
& & \multicolumn{3}{c}{ImageNet-1k} & \multicolumn{3}{c}{CIFAR-100} \\
\cmidrule(lr){3-5}\cmidrule(lr){6-8}
Input & params & reported & held out & infl. & reported & held out & infl. \\
\midrule
none (VarRatio)         & $0$                & 89.13 & 89.13 & $+0.00$ & 80.84 & 80.84 & $+0.00$ \\
$\{d_{(1)}, \operatorname{Var}\}$, target tuned & $1$ & 89.91 & 89.39 & $+0.52$ & 82.28 & 80.75 & $+1.53$ \\
2 channels              & $2$                & 89.22 & 88.47 & $+0.75$ & 81.30 & 79.32 & $+1.98$ \\
4 channels              & $4$                & 89.51 & 88.30 & $+1.21$ & 82.16 & 79.64 & $+2.52$ \\
8 channels              & $8$                & 91.69 & 90.04 & $+1.66$ & 84.93 & 80.46 & $+4.47$ \\
13 channels             & $13$               & 92.09 & 90.15 & $+1.94$ & 85.23 & 80.42 & $+4.80$ \\
\midrule
$\{d_{(1)}, \operatorname{Var}\}$ ridge & $2$ & 90.02 & 89.35 & $+0.67$ & 82.28 & 80.60 & $+1.69$ \\
13 channels, MLP        & $1.4\cdot 10^{5}$  & 93.37 & 89.84 & $+3.53$ & 87.62 & 79.77 & $+7.85$ \\
$z$, ridge              & $D$                & 98.91 & 91.65 & $+7.26$ & 94.51 & 74.70 & $+19.82$ \\
$z$, MLP                & $6.6\cdot 10^{5}$  & 99.08 & 91.41 & $+7.68$ & 94.21 & 74.54 & $+19.67$ \\
\midrule
\multicolumn{8}{l}{\textit{All five rows read 13 directions and hold 13 parameters}} \\[1pt]
sorted profile          & $13$               & 90.79 & 88.52 & $+2.27$ & 85.35 & 79.70 & $+5.65$ \\
unsorted profile        & $13$               & 75.13 & 63.68 & $+11.45$ & 82.44 & 64.61 & $+17.83$ \\[3pt]
13 channels             & $13$               & 92.09 & 90.15 & $+1.94$ & 85.23 & 80.42 & $+4.80$ \\
13 random views of $z$  & $13$               & 69.35 & 58.44 & $+10.92$ & 75.11 & 54.70 & $+20.41$ \\
13 top PCA views of $z$ & $13$               & 74.71 & 62.26 & $+12.45$ & 81.11 & 61.15 & $+19.96$ \\
\midrule
$\{d_{(1)}, \operatorname{Var}\}$, validation tuned & $1$ & 89.40 & 89.41 & $-0.02$ & 80.74 & 80.73 & $+0.01$ \\
\bottomrule
\end{tabular}
\end{table}

\subsection{Class identity is what the fit buys}
\label{sec:exp-perm}

We read the first control by its inflation, which is the quantity we measure everywhere else, and Figure~\ref{fig:material} in the appendix sweeps the pair across width. The class aware member inflates by $8.33$ AUROC on ImageNet and by $16.83$ on CIFAR-100. The permutation invariant member inflates by $3.11$ and by $6.83$. The gap is $5.22$ and $10.00$, and the class aware member inflates more on 11 of 11 ImageNet backbones and on 6 of 6 CIFAR-100 architectures, which a sign test puts at $p = 0.001$ and $p = 0.031$.

Capacity cannot produce that difference, because both members hold the same $C + 1$ numbers in the same dimension and take the same fitting rule. The amount of information does not produce it either. Widening the sorted member from 13 directions to its full $1001$ adds $0.84$ to its inflation, while exposing class identity at a fixed 13 adds $9.18$. Class identity is what remains, so it is what the within protocol rewards and the hold-out protocol removes. Appendix~\ref{app:perm} crosses the two hold-out constructions with the two ridge rules of Appendix~\ref{app:protocol}, and the class aware member inflates more in all four cells, unanimously across backbones each time. The ordering of the two members under the hold-out protocol is the less stable reading, which is why we quote the inflation rather than the ordering. Appendix~\ref{app:ksweep} varies how many OOD datasets the hold-out arm may read, and the angular distance between the two members holds at every size.

\begin{figure}[t]
\centering
\includegraphics[width=\textwidth]{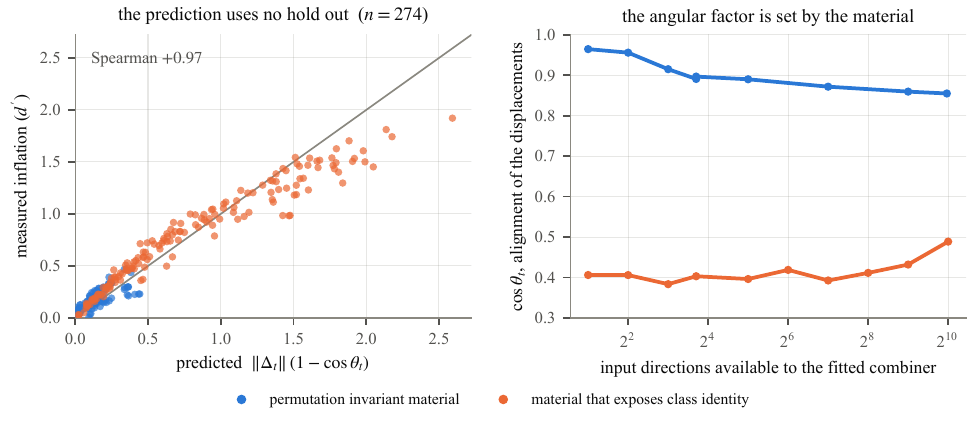}
\caption{The inflation is the product of a length and an angle. Left, the prediction of \eqref{eq:mech} against the inflation measured under the two protocols, one dot per representation per backbone. The prediction reads the fitting rows only. Right, the angular factor against the width of the fit. The factor separates the two materials by a wide margin at every width, which is why sorting the class distances flattens the width curve.}
\label{fig:mech}
\end{figure}

\subsection{The material is what sets the inflation}
\label{sec:exp-material}

The material block of Table~\ref{tab:capacity} holds width and parameter count fixed at 13 and varies what the fit reads. Its first pair holds the content fixed as well. The sorted and unsorted profiles carry the same numbers in the same dimension under the same rule, and the class aware member inflates by much the larger amount on both benchmarks. Appendix~\ref{app:fpr} reads every row of Table~\ref{tab:capacity} at the operating point instead, and the ordering is the same there. It inflates more on every one of our 17 backbones, which a sign test puts at $p = 1.5 \cdot 10^{-5}$. The second pair exchanges the channel inventory for a linear view of the whitened feature, and that view also inflates by much the larger amount on both benchmarks, again on every one of the 17. Leading principal directions inflate more than random ones on ImageNet and by a comparable amount on CIFAR-100, so the effect does not come from randomness in the projection. Figure~\ref{fig:width} in the appendix sweeps both materials across width, and the invariant material saturates while the class aware one climbs until the AUROC scale caps it. Appendix~\ref{app:perset} resolves these rows by OOD dataset, and the inflation is positive on every dataset of both benchmarks rather than carried by one of them.

\subsection{The inflation is a length times an angle}
\label{sec:exp-mech}
\label{sec:mech}

The two controls say what the inflation responds to. A linear score has a closed form that says why. Write $\Delta_t = \mu_t - \mu_{\text{ID}}$ for the displacement of OOD set $t$ from the in-distribution data, and take norms and angles in the $\Sigma^{-1}$ inner product that the discriminant works in. A score $w$ separates the two clouds at $d' = w^{\top}\Delta_t / \lVert w \rVert_{\Sigma}$. The within protocol hands the fit $w = \Sigma^{-1}\Delta_t$ and the hold-out protocol hands it $w = \Sigma^{-1}\bar{\Delta}$, so the two reach $\lVert \Delta_t \rVert$ and $\langle \bar{\Delta}, \Delta_t \rangle / \lVert \bar{\Delta} \rVert$ and their difference is
\begin{equation}
d'_{\text{within}} - d'_{\text{hold out}} \;=\; \lVert \Delta_t \rVert \,\big(1 - \cos \theta_t\big),
\label{eq:mech}
\end{equation}
where $\bar{\Delta}$ is the mean displacement of the other datasets, weighted by how many rows each one contributes, and $\theta_t$ is the angle between $\Delta_t$ and it. That weighting is what pooling the rows hands the hold-out arm, and reading both arms in one shared metric is what lets the gap factor into a length and an angle. Both factors come from the fitting rows. Neither needs a hold-out pass.

The two factors answer to different things. The norm grows as the fit gains directions, because the channel widths are nested and a random view of width $m$ carries $m/D$ of the squared length in expectation. The cosine is scale free and is what a permutation invariant readout collapses. Such a readout sees the class distances without seeing which class produced which one, so two OOD datasets that sit near different classes in the same relative position hand it the same numbers and their displacements come out parallel. Width and material are therefore the two factors of a product rather than two explanations competing for one slot.

We read \eqref{eq:mech} on the binormal scale rather than in AUROC. AUROC compresses near 1, and the within arm of the widest fits reaches $98.91$ while the hold-out arm still has room, so an AUROC gap understates what happens at the top of the range. We report AUROC everywhere else because that is what the literature reports and what a practitioner reads. We use $d' = \sqrt{2}\,\Phi^{-1}(\text{AUROC})$ only where a mechanism is at stake.

Figure~\ref{fig:mech} places that prediction against the measurement for every representation on every one of the 11 ImageNet backbones. That is 25 representations on each backbone, less one on ResNet-18 where $D = 512$ makes the 512 direction view a rotation of the whole space and so a copy of $z$, which leaves $274$ pairs. The rank correlation over them is $0.973$ and the Pearson correlation is $0.975$. The rank correlation reaches $0.844$ inside the permutation invariant material and $0.984$ inside the class aware material, and its median inside a single backbone is $0.982$. A line through the origin fits at slope $0.90$, so the closed form is a point prediction and not a ranking alone. What it gives is the gap between two population optima, and a regularized fit on finite rows lands a little under that at the widest fits and a little over it at the narrowest.

The measurement separates the two factors cleanly. Permutation invariant material sits at a mean cosine of $0.910$ and class aware material at $0.418$ when averaged over every representation, and that separation holds at every width. Width moves the class aware cosine from $0.41$ at 2 directions to $0.49$ at full width, so alignment improves a little as the fit widens. The length moves the other way and by far more. Random views of the feature carry $0.16$ at 2 directions and $3.55$ at full width. The sorted profile carries $2.41$ at 13 directions and $2.77$ at its full $1001$, so it saturates almost at once, which is the same saturation that the inflation shows.

One comparison makes the split concrete. At 13 directions and 13 parameters the invariant channel inventory has a displacement of length $1.61$ and the random view of the feature has one of length $0.55$. The invariant material carries the larger displacement by a factor of $2.9$ and inflates less, because the invariant cosine is $0.891$ against $0.403$. What the class aware fit gains is not a longer displacement but a displacement that points somewhere that the other datasets do not.

\section{Limitations and future work}

Four things bound what we measured. Every combiner we fit is a ridge discriminant or one network, so another fitting rule could place the split somewhere else. The closed form of Section~\ref{sec:mech} assumes a linear score under one covariance shared by the in-distribution and OOD clouds, so it describes the ridge discriminants exactly and the network only by analogy, and it says what the inflation is rather than what makes one material expose more class identity than another. That form gives the gap between two population optima, and a fit on finite rows lands near but not on it. The measured gap runs from three quarters to five thirds of the predicted one across the sweep, falling short of the prediction at the widest fits and running past it at the narrowest. The ImageNet backbones are public checkpoints whose seed we cannot vary, so the seed control runs on CIFAR-100 alone. The transfer result rests on two benchmarks and a third would sharpen it. One route past the constant also stays open. Choosing it from in-distribution data alone would remove the last reason to touch OOD data, and we tested the strongest construction that the literature offers, which retrains the classifier on subsets of its own classes \citep{mondal2026tuning}. On CIFAR-100 it does not reach the zero capacity reference and the value that it picks swings by an order of magnitude between class splits. Whether a surrogate exists that carries the kind of novelty that the constant responds to is the question that we leave open.

\section{Conclusion}

A fitted OOD score reports a number that its own validation cannot certify. Splitting that number into the part which survives a change of OOD dataset and the part which does not shows that most of it is dataset identity. The parameters that a fit holds do not move what survives. Four orders of magnitude at a fixed input raise the reported column and leave the held-out column where it was, while one change of material at a fixed parameter count moves both. The share depends instead on whether the input exposes class identity, and two controls that vary that property alone separate the inflation on every backbone of both benchmarks. A closed form explains both. The gap is a displacement length times $1 - \cos\theta$, and width sets the first factor while what the input exposes sets the second. Their product predicts the measurement at a rank correlation of $0.973$ without running the hold-out protocol at all. Across the capacity sequence the reported column climbs on both benchmarks while the column that survives climbs a little on ImageNet and falls on CIFAR-100. Anyone who fits on OOD data can run the same split with one extra pass.

\bibliography{iclr2027_conference}
\bibliographystyle{iclr2027_conference}

%

\clearpage

\appendix

\section{Reproducibility}
\label{app:repro}

\paragraph{Hardware and software.}
Every experiment runs on one NVIDIA RTX 4090 under WSL2 on Windows. We use PyTorch for the forward passes and for every fit, and \texttt{timm} \citep{wightman2019timm} for the ImageNet checkpoints that torchvision does not carry. All distance computations and all covariance estimates run in float64 on the GPU. We cache the penultimate features once per backbone in float16 and every later experiment reads that cache, so no result depends on a second forward pass.

\paragraph{Datasets.}
ImageNet-1k \citep{russakovsky2015imagenet} supplies 1{,}281{,}167 training images and 50{,}000 validation images. The OOD side follows OpenOOD v1.5 and holds \ds{SSB-hard} with 49{,}000 images \citep{vaze2022ssb}, \ds{NINCO} with 5{,}878 \citep{bitterwolf2023ninco}, \ds{iNaturalist} with 10{,}000 \citep{vanhorn2018inaturalist}, \ds{Textures} with 5{,}640 \citep{cimpoi2014dtd} and \ds{OpenImage-O} with 17{,}632 \citep{krasin2017openimages}. CIFAR-100 \citep{krizhevsky2009cifar} supplies 50{,}000 training and 10{,}000 test images. Its OOD side holds \ds{CIFAR-10} with 10{,}000 \citep{krizhevsky2009cifar}, \ds{TIN} with 10{,}000 \citep{le2015tinyimagenet}, \ds{MNIST} with 10{,}000 \citep{lecun1998mnist}, \ds{SVHN} with 26{,}032 \citep{netzer2011svhn}, \ds{Textures} with 5{,}640 and \ds{Places365} with 10{,}000 \citep{zhou2018places}. OpenOOD v1.5 fixes the near and far grouping we report. \ds{SSB-hard} and \ds{NINCO} are near for ImageNet and the other three are far, and \ds{CIFAR-10} and \ds{TIN} are near for CIFAR-100 and the other four are far. OpenOOD also designates one of those OOD datasets as a validation split, which is the split that the validation tuned condition of Section~\ref{sec:protocol} draws on, and Appendix~\ref{app:protocol} names the one it picks for each benchmark.

\paragraph{Metrics.}
We report AUROC throughout and FPR@95 beside it wherever we compare detectors, with the convention that a larger score means more out of distribution. The inflation is a difference of two AUROC values and we report it in those units throughout, with Appendix~\ref{app:fpr} giving the same rows at the operating point. FPR@95 is then the fraction of OOD samples that fall at or below the 95th percentile of the ID score, so a lower value is better. An average over a group of datasets is the plain mean over its members and an overall average is the plain mean over all of them. That choice matters. Averaging the near group and the far group first weights each group equally rather than each dataset and moves CIFAR-100 numbers by more than one AUROC.

\paragraph{Three deviations on the CIFAR side.}
Our \ds{TIN} split is the original TinyImageNet validation set of 10{,}000 images and not the deduplicated subset that OpenOOD provides, which makes it slightly nearer to the ID data. Our \ds{Places365} split is a fixed seed 0 draw of 10{,}000 images from the \texttt{val\_256} partition. Every CIFAR OOD image passes through a resize to 32 pixels followed by a center crop of 32 pixels.

\paragraph{ImageNet backbones.}
We use 11 public checkpoints and never train one. The torchvision side holds ResNet-18 with $D = 512$, ResNet-50 V1 and ResNet-50 V2 with $D = 2048$, ViT-B/16 and ViT-B/16 SWAG with $D = 768$, Swin-B and ConvNeXt-B with $D = 1024$, and MobileNetV3-Large with $D = 1280$. The rest come from \texttt{timm} and match the checkpoints of Mahalanobis++, namely ConvNeXtV2-B and SwinV2-B with $D = 1024$ and DeiT3-B/16 with $D = 768$. Each backbone keeps the preprocessing that its own checkpoint declares.

\paragraph{CIFAR-100 backbones.}
No pretrained CIFAR-100 checkpoint exists in \texttt{timm}, so we train our own under one frozen recipe. The recipe never varies with the architecture, because a per architecture recipe would add a degree of freedom that we cannot audit. We use SGD with learning rate $0.1$, momentum $0.9$ and Nesterov acceleration, weight decay $5 \cdot 10^{-4}$, batch size 128, 200 epochs and a cosine schedule to zero. The loss is plain cross entropy with no label smoothing and no mixup, since both reshape the feature geometry that this paper measures. Augmentation is a random crop of 32 pixels with padding 4 followed by a horizontal flip. Training runs under bfloat16 autocast.

\paragraph{Seeds.}
We train each of the 6 architectures three times under seeds 0, 1 and 2. Architecture is the unit of every count in the paper and seed is a repeat inside it, so a claim that holds on 6 of 6 backbones holds on 6 architectures and not on 18 draws. Table~\ref{tab:cifar-backbones} lists every run. The spread across seeds reaches $0.56$ points at most and the ordering of the architectures holds under all three seeds.

\begin{table}[t]
\centering
\caption{CIFAR-100 backbones. Top-1 accuracy for each seed, feature width $D$, parameter count and training time on one RTX 4090. The recipe is identical for every row and only the seed changes.}
\label{tab:cifar-backbones}
\small
\begin{tabular}{lrrrrrrr}
\toprule
Architecture & $D$ & Params & seed 0 & seed 1 & seed 2 & mean $\pm$ sd & minutes \\
\midrule
ResNet-18      &  512 & 11.2M & 78.40 & 78.15 & 77.60 & 78.05 $\pm$ 0.41 & 8 \\
ResNet-34      &  512 & 21.3M & 78.24 & 78.44 & 79.11 & 78.60 $\pm$ 0.46 & 15 \\
WRN-28-10      &  640 & 36.5M & 81.64 & 81.58 & 81.71 & 81.64 $\pm$ 0.07 & 57 \\
DenseNet-BC    &  342 & 0.80M & 77.92 & 77.05 & 77.13 & 77.37 $\pm$ 0.48 & 38 \\
VGG16-BN       &  512 & 14.8M & 74.85 & 74.45 & 74.46 & 74.59 $\pm$ 0.23 & 6 \\
MobileNetV2    & 1280 & 2.35M & 72.74 & 71.64 & 71.98 & 72.12 $\pm$ 0.56 & 14 \\
\bottomrule
\end{tabular}
\end{table}

\paragraph{Fitting the class statistics.}
We estimate $\mu_c$ and the tied $\Sigma$ on the full training split of the ID dataset after normalizing every feature to unit length. We shrink the covariance toward a scaled identity with strength $10^{-2}$ before inverting it. Appendix~\ref{app:units} shows that this strength moves nothing that this paper measures.

\paragraph{Baselines.}
We implement every baseline ourselves and run all of them through one protocol, so no number in this paper mixes our pipeline with a published one. Four baselines carry a constant that their papers fix by a rule. We give each of them an oracle value instead. We pick that value per backbone on the very test sets that we then score, over $D \in \{16, \dots, 1024\}$ for ViM, $k \in \{1, \dots, 1000\}$ for KNN and $\gamma \in \{0.01, \dots, 1.0\}$ for GEN and a clipping percentile in $\{80, 85, 90, 95, 99, 99.5, 99.9, 99.99\}$ for ReAct. A baseline that still loses under that handicap loses under any protocol.

\paragraph{Baseline calibration.}
Our numbers have to line up with published ones before any comparison in this paper means something. OpenOOD v1.5 reports six of our baselines on torchvision ResNet-50 V1, which is the one backbone that both sides run. Table~\ref{tab:openood} puts the two side by side. Four methods agree to within $0.02$ AUROC on the near group. Three methods sit about $0.67$ low on the far group and the rest between $0.13$ and $1.30$, which points at the composition of the far sets rather than at any one detector. KL-Matching sits $1.3$ to $1.5$ low because we estimate its templates on the training split while OpenOOD estimates them on the validation split. We do not compare against the OpenOOD entry for Mahalanobis, since that entry runs the original form without the length normalization that Mahalanobis++ adds.

\begin{table}[t]
\centering
\caption{Our implementation against OpenOOD v1.5 on ImageNet-1k with torchvision ResNet-50 V1. AUROC in percent, grouped as OpenOOD groups them.}
\label{tab:openood}
\small
\begin{tabular}{lcccccc}
\toprule
& \multicolumn{3}{c}{near} & \multicolumn{3}{c}{far} \\
\cmidrule(lr){2-4}\cmidrule(lr){5-7}
Method & ours & OpenOOD & $\Delta$ & ours & OpenOOD & $\Delta$ \\
\midrule
MSP         & 76.01 & 76.02 & $-0.01$ & 84.56 & 85.23 & $-0.67$ \\
MaxLogit    & 76.44 & 76.46 & $-0.02$ & 88.89 & 89.57 & $-0.68$ \\
Energy      & 75.88 & 75.89 & $-0.01$ & 88.79 & 89.47 & $-0.68$ \\
ViM         & 72.07 & 72.08 & $-0.01$ & 92.55 & 92.68 & $-0.13$ \\
KNN         & 71.48 & 71.10 & $+0.38$ & 89.71 & 90.18 & $-0.47$ \\
KL-Matching & 75.16 & 76.64 & $-1.48$ & 86.30 & 87.60 & $-1.30$ \\
\bottomrule
\end{tabular}
\end{table}

\section{The two protocols and the combiners}
\label{app:protocol}

Every measurement in Section~\ref{sec:protocol} runs on one set of splits. We cut the ID evaluation split into halves $A$ and $B$ under a fixed permutation with seed 0, and we cut each OOD set $t$ into halves $t_a$ and $t_b$ under seed 1. No fit half ever appears in an evaluation. Both protocols score the same rows $(B, t_b)$ for the same $t$, so \eqref{eq:within} and \eqref{eq:holdout} differ in what the fit reads and in nothing else, which is what makes their difference readable as one quantity.

Cutting the OOD set in half matters. Take a rule that reads a whole OOD set and then scores the same images. It measures how well it memorized those images. The within protocol splits the set first, so the fit sees the dataset without seeing the rows that it will be scored on. That is the protocol that a paper runs when it selects a constant on OOD validation data, and it is the one we want to price.

The sequence uses three fitting rules. The single constant comes from a grid of 41 values, namely $0$ together with a geometric grid from $10^{-4}$ to $10$. The within condition takes the value that maximizes AUROC on the fit half of the same set and the hold-out condition takes the value that maximizes the mean over the fit halves of the other sets. The ridge discriminants take a ridge strength from 7 values between $10^{-6}$ and $10$, chosen on a nested split inside the fit half. The evaluation half never enters a selection at any capacity point. The network holds two hidden layers of 512 and 256 units with GELU activations and dropout $0.1$, trains under AdamW at learning rate $10^{-3}$ and weight decay $10^{-4}$ with batch size 512 for at most 60 epochs, and stops early after 10 epochs without improvement on the nested split. On 17 inputs it holds about $1.4 \cdot 10^{5}$ parameters.

The validation tuned condition of Section~\ref{sec:protocol} runs on its own grid of 26 values, namely $0$ together with a geometric grid from $10^{-3}$ to $10$, which is the grid that the paper we follow reports \citep{kim2026mahavar}. It picks the constant on the fit half of the designated validation dataset and then scores every test set, and its hold-out arm picks on the fit halves of the other test sets. Both arms come from the same run, so their difference is internally consistent even though the grid differs from the 41 values that the capacity sequence uses. The two grids agree to $0.02$ AUROC at the hold-out arm.

OpenOOD designates \ds{OpenImage-O} for ImageNet and \ds{TIN} for CIFAR-100, and both of those are themselves reported test sets in the same benchmark. The validation tuned condition is therefore a within condition on one of the five ImageNet sets and on one of the six CIFAR-100 sets, and a hold-out condition on the rest. The target tuned row of Table~\ref{tab:capacity} is the pure within number at the same capacity.

Every experiment in this paper fits one rule on the union of the other sets' fit halves, so the hold-out arm reads the rows themselves rather than a combination of separate fits. Pooling weights each dataset by how many rows it brings, and the suites are uneven enough for that to matter, since \ds{SSB-hard} carries $55.6$ percent of the ImageNet OOD rows and \ds{SVHN} carries $36.3$ percent on CIFAR-100. An alternative normalizes the weight vector of each other set and sums those, which gives every dataset the same say. Table~\ref{tab:permcells} crosses the two and the reading of Section~\ref{sec:exp-perm} holds under both, so the uneven sizes do not carry it. We use the union everywhere because it applies to any fitting rule, while the normalized sum needs a single weight vector and so cannot reach the two network steps of the sequence.

The ridge strength is a second choice. Both experiments pick it on a nested split inside the fit half, so the discriminant sees half of the fit rows rather than all of them at one fixed value. Section~\ref{sec:exp-perm} reports the permutation control under that rule and Appendix~\ref{app:perm} gives all four combinations of the two choices.

\section{The channel inventory}
\label{app:channels}

The capacity sequence of Section~\ref{sec:sequence} fits ridge discriminants on prefixes of one fixed list of 17 scalar channels. We built that list for an earlier question and we did not reorder it for this paper, so the prefixes $m \in \{2, 4, 8, 17\}$ nest by construction. Every channel goes through the empirical CDF of the ID training split and then through $\Phi^{-1}$, which puts each one exactly at $\mathcal{N}(0,1)$ on ID data and removes any scale choice from the sum.

Table~\ref{tab:channels} gives the list. Write $d_{(1)} \le d_{(2)}$ for the two smallest class distances, $\hat{c}$ for the class that the classifier predicts, $L_c$ for the classifier logits and $L_{(1)} \ge L_{(2)}$ for the two largest of those. A spread in the table is a standard deviation.

\begin{table}[t]
\centering
\caption{The 17 channels in cache order. The last column marks the ones that CIFAR-100 does not define in our setup, which leaves the 13 channels that both benchmarks share. Channels 16 and 17 standardize channels 9 and 10 against the ID training split and then take the absolute value, so they answer how unusual a logit gap is rather than how large.}
\label{tab:channels}
\small
\begin{tabular}{clll}
\toprule
\# & Channel & Definition & Both \\
\midrule
1  & nearest distance    & $\log d_{(1)}$ & \checkmark \\
2  & whitened norm       & $\lVert z \rVert^{2}$ & \checkmark \\
3  & two nearest         & $\log d_{(1)} + \log d_{(2)}$ & \checkmark \\
4  & soft nearest        & $-\log \sum_c e^{-d_c/2}$ & \checkmark \\
5  & angular residual    & $1 - \max_c \cos(\hat{\phi}, \Sigma^{1/2}\mu_c)$ & \checkmark \\
6  & distance variance   & $-\operatorname{Var}_c d_c$ & \checkmark \\
7  & free energy         & $-\log \sum_c e^{L_c}$ & \checkmark \\
8  & relative gap        & $-\log\!\big((d_{(2)} - d_{(1)})/d_{(1)}\big)$ & \checkmark \\
9  & centered top logit  & $-\big(L_{(1)} - \operatorname*{mean}_c L_c\big)$ & \checkmark \\
10 & logit margin        & $-\big(L_{(1)} - L_{(2)}\big)$ & \checkmark \\
11 & prediction gap      & $\max\!\big(d_{\hat{c}} - d_{(1)}, \, 0\big)$ & \checkmark \\
12 & eigenbasis spread   & spread of the residual to $\mu_{\hat{c}}$ over covariance eigencoordinates & \\
13 & neighbor distance   & whitened distance to the nearest training feature & \\
14 & neighbor agreement  & whether that neighbor carries the label $\hat{c}$ & \\
15 & layer typicality    & spread of the standardized per layer activation rates & \\
16 & top logit deviation & $\big|\,$channel 9, standardized$\,\big|$ & \checkmark \\
17 & margin deviation    & $\big|\,$channel 10, standardized$\,\big|$ & \checkmark \\
\bottomrule
\end{tabular}
\end{table}

The four channels that CIFAR-100 does not carry are the eigenbasis spread, the two nearest neighbor terms and the layer typicality spread. When we run the sequence on both benchmarks together we drop those four and use the 13 that both sides define the same way. The prefixes at $m \in \{2, 4, 8\}$ are identical either way, so only the widest linear point changes between the two runs. The body quotes the 13 channel run throughout so that the two benchmarks sit on one axis. The ImageNet run on the native 17 channels puts the same steps in the same order and moves only the two points that involve the dropped channels. Its widest linear step reaches $92.97$ under the within protocol and $90.54$ under the hold-out protocol for an inflation of $2.43$, against $92.09$ and $90.15$ for an inflation of $1.94$ at 13 channels. Its network reaches $94.29$ and $90.45$ for an inflation of $3.84$, against $93.37$ and $89.84$ for an inflation of $3.53$ at 13 channels. The whitened feature space point is the same run in both cases, because it reads $z$ rather than the inventory.

\section{The projection controls}
\label{app:proj}

Both controls of Section~\ref{sec:control} read their material through the same construction. For a material of width $D$ and a target width $m \le D$ we draw a $D \times m$ Gaussian matrix under a fixed seed, take the $Q$ factor of its $QR$ decomposition, and fit the ridge discriminant on the projected rows. The columns are orthonormal, so at $m = D$ the map is a rotation and the step reproduces a ridge on the material itself. On ImageNet that anchor holds to the second decimal against the whitened feature step of the capacity sequence, at $98.91$ within and $91.65$ held out in both, which is what licenses reading the sweep as one curve with the sequence. We report three seeds at every width below $D$ and average them; the seed standard deviation at 13 directions runs from $0.9$ to $2.9$ AUROC against effects of $6$ to $16$.

The projected combiners are poor detectors, and one reading of Section~\ref{sec:exp-material} is that poor detectors simply have noisy gaps. Two matched pairs close that reading. On CIFAR-100 the invariant material at 4 directions and the class aware material at full width sit within $0.02$ AUROC of each other against the zero capacity reference, at $-4.74$ and $-4.72$, and they inflate $4.14$ and $16.83$. On ImageNet the class aware material at full width is $1.59$ AUROC better than the invariant material at 13 directions and inflates $3.7$ times as much. The better detector carries the larger gap.

The first control projects the two profile views, both of width $C+1$. The second projects the whitened feature $z$, whose width is the feature width of the backbone. We also report the leading principal directions of $z$ in place of random ones, with the basis estimated on the in-distribution fit half alone so that no evaluation row enters it. Splits, ridge selection and the two protocols are the ones of Appendix~\ref{app:protocol}, unchanged.

CIFAR-100 carries no whitened feature step in the capacity sequence, because the four channels that it does not define are the ones that set the widest linear point. The projection sweep supplies it. At $m = D$ the step is a ridge on the whitened feature, and it reaches $94.51$ within and $74.70$ held out for an inflation of $19.82$. That value is the CIFAR-100 entry of the widest row of Table~\ref{tab:capacity}.

\section{The reference against the field}
\label{app:ref}

\begin{figure}[t]
\centering
\includegraphics[width=\textwidth]{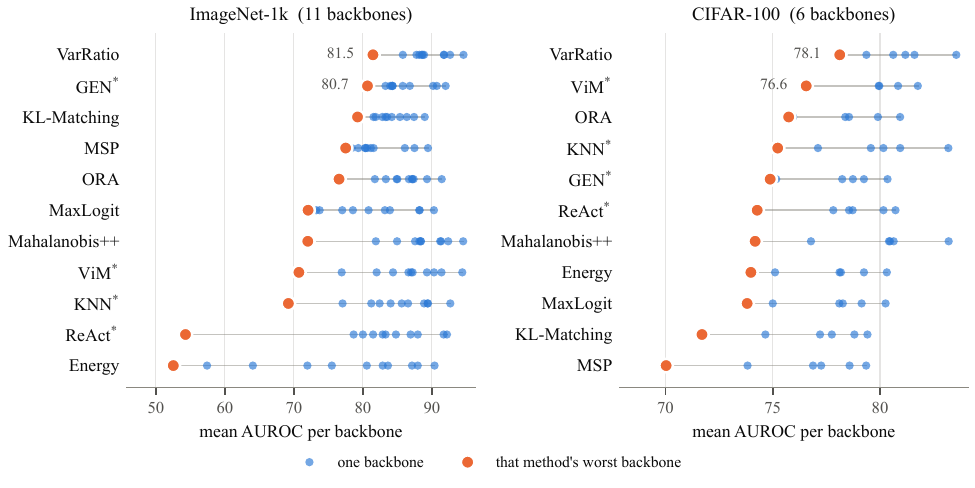}
\caption{Detectors behave differently on different backbones. Each dot is one backbone and the highlighted dot is that detector's worst backbone. Rows sort by that worst case. A star marks a detector whose constant we give an oracle value on the test sets.}
\label{fig:worst}
\end{figure}

Section~\ref{sec:exp-ref} states three checks on the reference and this section carries them. Table~\ref{tab:main} places VarRatio against 10 detectors on both benchmarks, where it leads the average AUROC and the average FPR@95 on both and leads the near and the far group as well on ImageNet. Against Mahalanobis++ it improves the average AUROC on 11 of 11 ImageNet backbones with a minimum margin of $0.05$ and on 5 of 6 CIFAR-100 architectures, and FPR@95 on 11 of 11 and on 4 of 6. Figure~\ref{fig:worst} gives the spread. The third check fits a ridge discriminant on the two quantities that VarRatio reads, over four OOD datasets, and scores it on the fifth. It reaches $89.35$ against $89.13$ on ImageNet, so the reference sits within $0.22$ of the honest ceiling of its own material. Appendix~\ref{app:full} resolves every detector by backbone and by OOD dataset. Detectors vary enormously across backbones, and Energy alone spans $52.5$ to $90.4$ AUROC on ImageNet, so the worst case is what a practitioner without OOD data is exposed to.

\begin{table}[t]
\centering
\caption{Detectors that need no OOD data, averaged over backbones. AUROC and FPR@95 in percent. A star marks a detector whose constant we hand an oracle value on the test sets. CIFAR-100 rows average the 3 seeds of each architecture.}
\label{tab:main}
\small
\begin{tabular}{lcccccc}
\toprule
& \multicolumn{2}{c}{Near-OOD} & \multicolumn{2}{c}{Far-OOD} & \multicolumn{2}{c}{Avg} \\
\cmidrule(lr){2-3}\cmidrule(lr){4-5}\cmidrule(lr){6-7}
Method & AUROC & FPR & AUROC & FPR & AUROC & FPR \\
\midrule
\multicolumn{7}{l}{\textit{ImageNet-1k, 11 backbones}} \\[1pt]
MSP & 77.17 & 73.23 & 85.24 & 53.64 & 82.01 & 61.47 \\
MaxLogit & 75.54 & 70.83 & 84.36 & 50.31 & 80.83 & 58.52 \\
Energy & 71.75 & 73.35 & 78.52 & 56.73 & 75.81 & 63.38 \\
GEN$^{*}$ & 79.89 & 69.40 & 90.09 & 42.30 & 86.01 & 53.14 \\
KL-Matching & 77.31 & 73.34 & 88.53 & 50.43 & 84.04 & 59.60 \\
KNN$^{*}$ & 76.05 & 76.91 & 89.66 & 44.43 & 84.21 & 57.42 \\
ViM$^{*}$ & 76.86 & 72.67 & 91.20 & 38.43 & 85.47 & 52.12 \\
ReAct$^{*}$ & 75.74 & 72.07 & 86.49 & 49.03 & 82.19 & 58.24 \\
ORA & 76.74 & 74.59 & 91.34 & 41.98 & 85.50 & 55.03 \\
Mahalanobis++ & 79.93 & 64.03 & 92.28 & 34.35 & 87.34 & 46.22 \\
VarRatio & 81.98 & 60.38 & 93.82 & 29.46 & 89.08 & 41.83 \\
\midrule
\multicolumn{7}{l}{\textit{CIFAR-100, 6 architectures}} \\[1pt]
MSP & 78.06 & 78.93 & 74.95 & 83.61 & 75.99 & 82.05 \\
MaxLogit & 78.62 & 77.85 & 76.84 & 82.14 & 77.43 & 80.71 \\
Energy & 78.53 & 78.23 & 76.98 & 81.88 & 77.49 & 80.66 \\
GEN$^{*}$ & 78.94 & 77.99 & 77.19 & 81.96 & 77.77 & 80.63 \\
KL-Matching & 77.82 & 78.96 & 75.98 & 80.46 & 76.59 & 79.96 \\
KNN$^{*}$ & 78.00 & 79.18 & 80.05 & 74.18 & 79.37 & 75.85 \\
ViM$^{*}$ & 75.35 & 84.38 & 81.27 & 70.96 & 79.30 & 75.43 \\
ReAct$^{*}$ & 77.81 & 78.48 & 78.66 & 79.73 & 78.38 & 79.32 \\
ORA & 78.37 & 78.64 & 78.18 & 79.24 & 78.25 & 79.04 \\
Mahalanobis++ & 75.38 & 82.27 & 81.23 & 68.86 & 79.28 & 73.33 \\
VarRatio & 76.54 & 80.50 & 82.84 & 68.89 & 80.74 & 72.76 \\
\bottomrule
\end{tabular}
\end{table}

\begin{figure}[t]
\centering
\includegraphics[width=\textwidth]{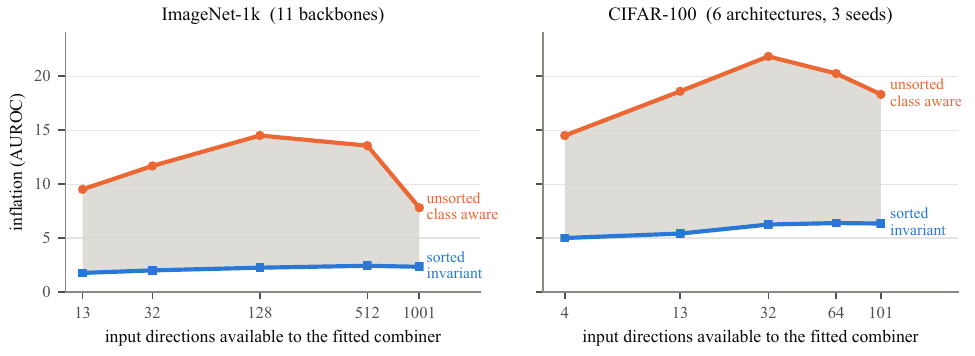}
\caption{The same numbers, sorted and unsorted. Both rows hold the $C+1$ entries of the class distance profile in the same dimension, take the same ridge discriminant under the same splits, and are read through the same random projections. Sorting the vector first removes the width effect and lowers the inflation by a factor of $2.9$ to $6.5$. The vertical distance between the two curves is what class identity costs.}
\label{fig:material}
\end{figure}

Figure~\ref{fig:material} belongs with Section~\ref{sec:exp-perm}. It sweeps the first control across width, and sorting the class distances first flattens the curve that the class aware member climbs.

\begin{figure}[t]
\centering
\includegraphics[width=\textwidth]{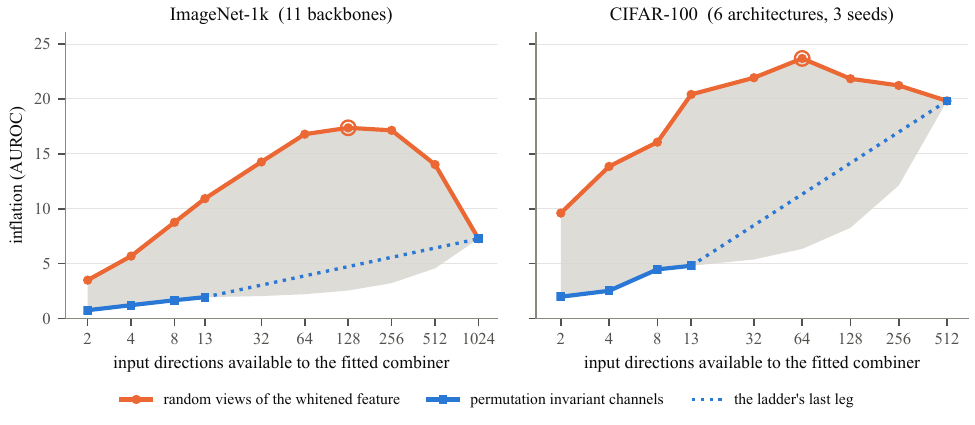}
\caption{The inflation against the width of the fit, on two materials. The orange line reads random orthonormal views of the whitened feature, which can respond to the distance of a named class. The blue line reads prefixes of the permutation invariant channel inventory. At any fixed point of the horizontal axis the two carry the same directions and the same free parameters. At the right edge they meet, because a full width projection is a rotation. The dotted leg is the last step of the capacity sequence, which changes the material and the width at once.}
\label{fig:width}
\end{figure}

Figure~\ref{fig:width} belongs with Section~\ref{sec:exp-material}. It sweeps the width of the fit on two materials at once, and the widest step of the capacity sequence moves along both axes together. The class aware curve peaks at $17.37$ on 128 directions and falls to $7.26$ once the fit may read all $1024$. That fall belongs to the AUROC scale and not to the fit. The within arm reaches $98.91$ and cannot climb further, while the hold-out arm still climbs from $71.38$ to $91.65$. On the binormal scale of Section~\ref{sec:mech} the same sweep runs $0.112$, $0.414$, $0.968$, $1.371$ and $1.443$ at 2, 13, 128, 512 and $1024$ directions, which is monotone and saturating.

The two materials separate on that scale as well. At 13 directions the invariant material sits at $0.17$ and the class aware material at $0.41$. Over a 77 fold change of width the invariant material moves from $0.18$ to $0.26$ while the class aware material moves from $0.38$ to $1.40$, so the invariant one saturates while the other keeps climbing.

\section{The permutation control by backbone}
\label{app:perm}

Two choices could carry this experiment, namely how the hold-out condition weights the other sets and how the ridge strength is chosen. Table~\ref{tab:permcells} crosses them, and the union row of each pair is the construction that every experiment in this paper uses. The class aware member inflates more than the permutation invariant one in all four cells and on every backbone of both benchmarks, so Section~\ref{sec:exp-perm} does not rest on either choice. The ordering of the two members under the hold-out protocol does rest on them. It reverses in all four cells on CIFAR-100 and in only two of them on ImageNet, which is why we quote the inflation instead.

\begin{table}[t]
\centering
\caption{The permutation pair under both hold-out constructions and both ridge rules. The inflation columns give the inflation of each member and the count of backbones on which the class aware member inflates more. The ordering columns subtract the hold-out AUROC of the permutation invariant member from that of the class aware one, and they give the count of backbones on which the class aware member is ahead. CIFAR-100 counts run over 6 architectures with 3 seeds averaged inside each.}
\label{tab:permcells}
\small
\begin{tabular}{llccccc}
\toprule
& & \multicolumn{3}{c}{inflation} & \multicolumn{2}{c}{ordering, held out} \\
\cmidrule(lr){3-5}\cmidrule(lr){6-7}
Hold out & Ridge & invariant & class aware & more & margin & ahead \\
\midrule
\multicolumn{7}{l}{\textit{ImageNet-1k, 11 backbones}} \\[1pt]
normalized sum & fixed & $+2.15$ & $+11.66$ & 11/11 & $-3.87$ & 2/11 \\
union & fixed & $+2.82$ & $+13.78$ & 11/11 & $-5.32$ & 2/11 \\
normalized sum & nested & $+2.33$ & $+7.81$ & 11/11 & $+0.92$ & 8/11 \\
union & nested & $+3.11$ & $+8.33$ & 11/11 & $+1.17$ & 9/11 \\
\midrule
\multicolumn{7}{l}{\textit{CIFAR-100, 6 architectures}} \\[1pt]
normalized sum & fixed & $+6.19$ & $+18.29$ & 6/6 & $-5.99$ & 0/6 \\
union & fixed & $+6.51$ & $+16.94$ & 6/6 & $-4.31$ & 1/6 \\
normalized sum & nested & $+6.34$ & $+18.32$ & 6/6 & $-5.88$ & 0/6 \\
union & nested & $+6.83$ & $+16.83$ & 6/6 & $-3.90$ & 2/6 \\
\bottomrule
\end{tabular}
\end{table}

Section~\ref{sec:exp-perm} reports the permutation pair averaged over backbones. Table~\ref{tab:permfull} resolves it. The class aware view inflates more on all 17 backbones and wins the within column on all 17. The class aware view loses the hold-out column on 6 of them, of which 4 are CIFAR-100 architectures. The size of the reversal moves much more than its sign. DenseNet-BC gains $4.63$ inside the dataset and loses $13.47$ once the dataset is held out, while VGG16-BN gains $7.66$ and holds its ground at $+0.27$.

\begin{table}[t]
\centering
\caption{The permutation pair at full width backbone by backbone, under the nested ridge rule of Appendix~\ref{app:protocol}. Sorted hides which class produced which distance and unsorted does not. Both views carry the same $C + 1$ numbers, take the same ridge discriminant and score the same rows, so only class identity separates them. AUROC in percent.}
\label{tab:permfull}
\small
\begin{tabular}{lcccccc}
\toprule
& \multicolumn{3}{c}{sorted, permutation invariant} & \multicolumn{3}{c}{unsorted, class aware} \\
\cmidrule(lr){2-4}\cmidrule(lr){5-7}
Backbone & within & held out & inflation & within & held out & inflation \\
\midrule
\multicolumn{7}{l}{\textit{ImageNet-1k, 11 backbones}} \\[1pt]
ConvNeXtV2-B     & 96.13 & 95.07 & $+1.06$ & 99.80 & 95.98 & $+3.82$ \\
SwinV2-B         & 94.83 & 92.31 & $+2.52$ & 99.80 & 94.49 & $+5.31$ \\
DeiT3-B/16       & 93.57 & 90.89 & $+2.69$ & 99.37 & 93.03 & $+6.33$ \\
ResNet-50 V1     & 91.55 & 88.74 & $+2.81$ & 96.82 & 84.54 & $+12.29$ \\
Swin-B           & 92.14 & 88.62 & $+3.52$ & 99.54 & 93.94 & $+5.61$ \\
ViT-B/16         & 90.32 & 88.36 & $+1.96$ & 98.52 & 89.05 & $+9.47$ \\
ResNet-50 V2     & 92.97 & 87.79 & $+5.18$ & 96.50 & 81.20 & $+15.30$ \\
ViT-B/16 SWAG    & 93.22 & 91.14 & $+2.09$ & 99.43 & 92.88 & $+6.55$ \\
ConvNeXt-B       & 91.61 & 89.38 & $+2.23$ & 99.60 & 94.11 & $+5.49$ \\
MobileNetV3-L    & 87.26 & 81.63 & $+5.63$ & 97.03 & 85.32 & $+11.71$ \\
ResNet-18        & 88.91 & 84.39 & $+4.51$ & 96.46 & 86.71 & $+9.75$ \\
\midrule
Mean             & 92.05 & 88.94 & +3.11 & 98.44 & 90.11 & +8.33 \\
\midrule
\multicolumn{7}{l}{\textit{CIFAR-100, 6 architectures, mean over 3 seeds}} \\[1pt]
\midrule
ResNet-18        & 87.81 & 81.89 & $+5.92$ & 93.75 & 79.97 & $+13.79$ \\
ResNet-34        & 86.05 & 81.25 & $+4.80$ & 93.34 & 78.27 & $+15.08$ \\
WRN-28-10        & 86.67 & 78.97 & $+7.70$ & 94.45 & 80.29 & $+14.16$ \\
DenseNet-BC      & 88.49 & 81.71 & $+6.78$ & 93.12 & 68.24 & $+24.89$ \\
VGG16-BN         & 84.89 & 77.42 & $+7.48$ & 92.55 & 77.68 & $+14.87$ \\
MobileNetV2      & 87.18 & 78.88 & $+8.30$ & 90.47 & 72.26 & $+18.21$ \\
\midrule
Mean             & 86.85 & 80.02 & +6.83 & 92.95 & 76.12 & +16.83 \\
\bottomrule
\end{tabular}
\end{table}

\section{The inflation at the operating point}
\label{app:fpr}

A deployed gate is set at a threshold, not at a ranking, so the number a practitioner lives with is the false positive rate at a fixed true positive rate. Every inflation in the body is a difference of two AUROC values. Table~\ref{tab:fpr} reads the same rows at FPR@95. Lower is better there, so the inflation is the held-out value minus the within one, and a positive entry means the reported gate lets through fewer OOD samples than the deployed one will. Both numbers come from the same fitted direction, so nothing but the metric changes between this table and Table~\ref{tab:capacity}.

The ordering survives the change of metric and the size does not. The widest fits separate far more at the operating point than in AUROC, because their within arm sits where the rate is already small. A ridge on the whitened feature reports a gate that misses $4.98$ percent of ImageNet OOD samples and delivers one that misses $35.32$, and on CIFAR-100 the same step reports $26.20$ and delivers $79.84$. The validation tuned row moves the other way on both benchmarks, which is the reading that Section~\ref{sec:exp-capacity} gives in AUROC. Ratios between rows do not carry across the two metrics, because FPR@95 compresses where AUROC does not and the rows sit at different operating points.

\begin{table}[t]
\centering
\caption{The inflation at FPR@95, over the 11 ImageNet backbones and the 6 CIFAR-100 architectures with their seeds averaged inside. Columns give the rate that the usual protocol reports, the rate that survives holding the OOD dataset out, and the difference. Lower is better, so a positive difference is an understated rate. Rows and splits are those of Table~\ref{tab:capacity}.}
\label{tab:fpr}
\small
\setlength{\tabcolsep}{4pt}
\begin{tabular}{lcccccc}
\toprule
& \multicolumn{3}{c}{ImageNet-1k} & \multicolumn{3}{c}{CIFAR-100} \\
\cmidrule(lr){2-4}\cmidrule(lr){5-7}
Input & reported & held out & infl. & reported & held out & infl. \\
\midrule
none (VarRatio)         & 41.70 & 41.70 & $+0.00$ & 72.72 & 72.72 & $+0.00$ \\
$\{d_{(1)}, \operatorname{Var}\}$, target tuned & 38.73 & 40.75 & $+2.03$ & 69.80 & 73.22 & $+3.42$ \\
2 channels              & 41.99 & 44.78 & $+2.79$ & 70.76 & 74.86 & $+4.10$ \\
4 channels              & 41.35 & 45.00 & $+3.65$ & 70.15 & 74.43 & $+4.27$ \\
8 channels              & 33.92 & 38.91 & $+4.98$ & 64.59 & 74.82 & $+10.23$ \\
13 channels             & 32.95 & 38.80 & $+5.85$ & 63.87 & 74.63 & $+10.76$ \\
\midrule
$\{d_{(1)}, \operatorname{Var}\}$ ridge & 38.69 & 40.53 & $+1.84$ & 70.04 & 73.53 & $+3.49$ \\
13 channels, MLP        & 29.27 & 39.06 & $+9.79$ & 55.97 & 75.51 & $+19.54$ \\
$z$, ridge              & 4.98 & 35.32 & $+30.33$ & 26.20 & 79.84 & $+53.64$ \\
$z$, MLP                & 4.17 & 32.45 & $+28.27$ & 26.59 & 77.83 & $+51.25$ \\
\midrule
\multicolumn{7}{l}{\textit{All five rows read 13 directions and hold 13 parameters}} \\[1pt]
sorted profile          & 36.34 & 42.27 & $+5.93$ & 62.34 & 75.04 & $+12.71$ \\
unsorted profile        & 75.15 & 87.01 & $+11.86$ & 61.52 & 87.08 & $+25.57$ \\[3pt]
13 channels             & 32.95 & 38.80 & $+5.85$ & 63.87 & 74.63 & $+10.76$ \\
13 random views of $z$  & 78.17 & 88.73 & $+10.55$ & 68.38 & 93.32 & $+24.94$ \\
13 top PCA views of $z$ & 81.17 & 94.29 & $+13.12$ & 65.83 & 94.18 & $+28.35$ \\
\midrule
$\{d_{(1)}, \operatorname{Var}\}$, validation tuned & 41.32 & 40.75 & $-0.56$ & 73.42 & 73.22 & $-0.21$ \\
\bottomrule
\end{tabular}
\end{table}

\section{The inflation dataset by dataset}
\label{app:perset}

Every inflation in this paper averages over the datasets that the benchmark supplies. With five or six of them one fold could carry the effect, and a paper about which OOD source a detector meets should say whether it does. Table~\ref{tab:perset} resolves the rows that the argument rests on. Each cell is the gap between the two protocols for one target dataset, averaged over backbones under the splits and the ridge rule of Appendix~\ref{app:protocol}, and each row average reproduces the matching row of Table~\ref{tab:capacity}.

No fold carries the effect. The inflation is positive in every cell of both benchmarks, and the largest single share of a row runs from $20$ to $42$ percent where an even split would give $20$ on ImageNet and $17$ on CIFAR-100. Which dataset inflates most does move with the material. The permutation invariant rows peak on \ds{SSB-hard}, the near set that sits closest to the in-distribution data, while the class aware rows at 13 directions peak on \ds{Textures} on ImageNet.

\begin{table}[t]
\centering
\caption{The inflation resolved by OOD dataset. Each cell gives $w(t) - h(t)$ for one target dataset in AUROC points, averaged over the 11 ImageNet backbones and over the 6 CIFAR-100 architectures with their seeds averaged inside. The last column is the mean over the datasets, which is the inflation that Table~\ref{tab:capacity} reports. The two profile rows read the control that Section~\ref{sec:exp-material} leads with at 13 directions, and the ImageNet block carries the channel inventory and the whitened feature as well.}
\label{tab:perset}
\small
\setlength{\tabcolsep}{4pt}
\begin{tabular}{lcccccc}
\toprule
\multicolumn{7}{l}{\textit{ImageNet-1k, 11 backbones}} \\[1pt]
Input & \ds{SSB-hard} & \ds{NINCO} & \ds{iNaturalist} & \ds{Textures} & \ds{OpenImage-O} & mean \\
\midrule
sorted profile           & $+4.03$ & $+0.65$ & $+1.00$ & $+3.59$ & $+2.08$ & $+2.27$ \\
unsorted profile         & $+13.21$ & $+7.29$ & $+10.83$ & $+16.67$ & $+9.26$ & $+11.45$ \\
13 channels              & $+3.57$ & $+0.50$ & $+0.63$ & $+3.49$ & $+1.48$ & $+1.94$ \\
13 random views of $z$   & $+9.57$ & $+6.40$ & $+15.24$ & $+17.07$ & $+6.30$ & $+10.92$ \\
$z$, ridge               & $+15.13$ & $+7.61$ & $+3.76$ & $+7.03$ & $+2.76$ & $+7.26$ \\
\bottomrule
\end{tabular}

\begin{tabular}{lccccccc}
\toprule
\multicolumn{8}{l}{\textit{CIFAR-100, 6 architectures}} \\[1pt]
Input & \ds{CIFAR-10} & \ds{TIN} & \ds{MNIST} & \ds{SVHN} & \ds{Textures} & \ds{Places365} & mean \\
\midrule
sorted profile           & $+5.53$ & $+2.95$ & $+11.04$ & $+5.75$ & $+3.20$ & $+5.41$ & $+5.65$ \\
unsorted profile         & $+20.91$ & $+11.33$ & $+18.79$ & $+20.01$ & $+17.07$ & $+18.85$ & $+17.83$ \\
\bottomrule
\end{tabular}
\end{table}

\section{The capacity sequence backbone by backbone}
\label{app:spread}

Table~\ref{tab:capacity} averages over backbones. Table~\ref{tab:spread} gives the spread behind every row of it. The inflation is positive on every ImageNet backbone from the second step on, the class aware material inflates more than the invariant one on all 11 ImageNet backbones and all 6 CIFAR-100 architectures, and the step from one parameter to 13 channels moves every backbone in the same direction on both benchmarks, at $p = 0.001$ over the 11 ImageNet backbones and $p = 0.031$ over the 6 CIFAR-100 architectures under a sign test. Both values are the smallest that a sign test can return at those counts, so they say that the direction is unanimous and nothing about how large the step is. On CIFAR-100 the seed standard deviation inside an architecture stays below the standard deviation across architectures at every step, which is why architecture is the unit of every count in this paper. The validation tuned row moves no ImageNet backbone by more than $0.28$ in either direction.

\begin{table}[t]
\centering
\caption{The spread behind the capacity sequence of Table~\ref{tab:capacity}. Inflation in AUROC points. The ImageNet columns run over 11 backbones and the CIFAR-100 columns over 6 architectures with the three seeds averaged inside each. The last column gives the seed standard deviation inside an architecture, averaged over the six.}
\label{tab:spread}
\small
\setlength{\tabcolsep}{4pt}
\begin{tabular}{lrrrrrrrrr}
\toprule
& \multicolumn{4}{c}{ImageNet-1k} & \multicolumn{5}{c}{CIFAR-100} \\
\cmidrule(lr){2-5}\cmidrule(lr){6-10}
Input & mean & sd & min & max & mean & sd & min & max & seed sd \\
\midrule
none (VarRatio)                            & $0.00$ & $0.00$ & $0.00$ & $0.00$ & $0.00$ & $0.00$ & $0.00$ & $0.00$ & $0.00$ \\
$\{d_{(1)}, \operatorname{Var}\}$          & $0.52$ & $0.55$ & $0.02$ & $1.97$ & $1.53$ & $1.14$ & $0.75$ & $3.66$ & $0.40$ \\
2 channels                                 & $0.75$ & $0.45$ & $0.09$ & $1.64$ & $1.98$ & $0.85$ & $0.56$ & $3.18$ & $0.67$ \\
4 channels                                 & $1.21$ & $0.72$ & $0.42$ & $2.85$ & $2.52$ & $1.47$ & $0.91$ & $5.24$ & $0.65$ \\
8 channels                                 & $1.66$ & $0.72$ & $0.60$ & $3.30$ & $4.47$ & $1.98$ & $2.53$ & $7.89$ & $0.89$ \\
13 channels                                & $1.94$ & $0.81$ & $0.76$ & $3.93$ & $4.80$ & $1.93$ & $2.92$ & $8.22$ & $0.85$ \\
13 channels, MLP                           & $3.53$ & $0.77$ & $2.05$ & $5.15$ & $7.85$ & $0.88$ & $6.72$ & $9.09$ & $1.06$ \\
$\{d_{(1)}, \operatorname{Var}\}$ ridge    & $0.67$ & $0.60$ & $0.04$ & $1.93$ & $1.69$ & $1.15$ & $0.75$ & $3.78$ & $0.47$ \\
13 random views of $z$                     & $10.92$ & $3.04$ & $6.53$ & $16.46$ & $20.41$ & $2.97$ & $17.56$ & $24.44$ & $3.42$ \\
$z$, ridge                                 & $7.26$ & $2.26$ & $4.49$ & $10.93$ & $19.82$ & $5.84$ & $15.26$ & $29.87$ & $1.55$ \\
$z$, MLP                                   & $7.68$ & $1.55$ & $6.07$ & $9.85$ & $19.67$ & $6.12$ & $14.00$ & $28.99$ & $1.73$ \\
13 top PCA views of $z$                    & $12.45$ & $3.34$ & $7.63$ & $19.62$ & $19.96$ & $2.47$ & $16.26$ & $22.97$ & $1.79$ \\
$\{d_{(1)}, \operatorname{Var}\}$, validation tuned & $-0.02$ & $0.16$ & $-0.28$ & $0.27$ & $0.01$ & $0.61$ & $-1.19$ & $0.52$ & $0.25$ \\
\bottomrule
\end{tabular}
\end{table}

\section{How many OOD datasets the fit may see}
\label{app:ksweep}

The hold-out arm of Section~\ref{sec:protocol} pools the fit rows of every other OOD dataset. Table~\ref{tab:ksweep} pools only $k$ of them and averages over every subset of that size, so the fitting suite grows from one dataset up to all that the benchmark offers. Both materials inflate less as $k$ grows and both are still falling at the largest $k$ the benchmark allows, so a larger fitting suite buys a real amount and these two benchmarks cannot say where it stops. The angular factor tells the same story from the other side. Both cosines rise with $k$ and the distance between them does not close, at $0.44$ to $0.42$ on ImageNet and $0.37$ to $0.41$ on CIFAR-100. The largest $k$ reproduces the cosines that Section~\ref{sec:exp-mech} reads, because both come from the same construction. A fit that reads class identity is therefore not misaligned because four datasets estimate the shared direction poorly. It is misaligned because it points at something the other datasets do not share, and adding datasets of the kind these benchmarks contain does not repair that.

\begin{table}[t]
\centering
\caption{The fitting suite as a variable. For each target dataset the hold-out arm reads $k$ of the others, averaged over every subset of that size. Held out AUROC and inflation in percent, and $\cos\theta$ between the direction the within arm gets and the one the hold-out arm gets. The within column does not depend on $k$ and sits at $92.05$ and $98.44$ on ImageNet and at $86.85$ and $92.95$ on CIFAR-100. At $k = 1$ the two constructions of Appendix~\ref{app:protocol} coincide, because one dataset has nothing to be combined with. The largest $k$ reproduces Section~\ref{sec:exp-perm}.}
\label{tab:ksweep}
\small
\begin{tabular}{lrrrrrrr}
\toprule
& \multicolumn{3}{c}{sorted, permutation invariant} & \multicolumn{3}{c}{unsorted, class aware} & \\
\cmidrule(lr){2-4}\cmidrule(lr){5-7}
$k$ & held out & inflation & $\cos\theta$ & held out & inflation & $\cos\theta$ & gap \\
\midrule
\multicolumn{8}{l}{\textit{ImageNet-1k, 11 backbones, 5 OOD datasets}} \\[1pt]
$1$ & $88.27$ & $+3.78$ & $0.796$ & $84.80$ & $+13.65$ & $0.355$ & $+9.87$ \\
$2$ & $88.72$ & $+3.33$ & $0.832$ & $87.90$ & $+10.55$ & $0.400$ & $+7.21$ \\
$3$ & $88.85$ & $+3.20$ & $0.847$ & $89.34$ & $+9.10$ & $0.423$ & $+5.91$ \\
$4$ & $88.94$ & $+3.11$ & $0.855$ & $90.11$ & $+8.33$ & $0.439$ & $+5.22$ \\
\midrule
\multicolumn{8}{l}{\textit{CIFAR-100, 6 architectures, 6 OOD datasets}} \\[1pt]
$1$ & $76.40$ & $+10.45$ & $0.615$ & $67.02$ & $+25.93$ & $0.240$ & $+15.48$ \\
$2$ & $78.37$ & $+8.48$ & $0.672$ & $70.56$ & $+22.39$ & $0.264$ & $+13.91$ \\
$3$ & $79.27$ & $+7.58$ & $0.699$ & $72.93$ & $+20.02$ & $0.283$ & $+12.44$ \\
$4$ & $79.73$ & $+7.12$ & $0.715$ & $74.59$ & $+18.36$ & $0.302$ & $+11.24$ \\
$5$ & $80.02$ & $+6.83$ & $0.726$ & $76.12$ & $+16.83$ & $0.321$ & $+10.00$ \\
\bottomrule
\end{tabular}
\end{table}

\section{A parameter axis at fixed input}
\label{app:mlpwidth}

The parameter block of Table~\ref{tab:capacity} raises the parameter count three times, and only one of those steps spans four orders of magnitude. Table~\ref{tab:mlpwidth} builds the axis directly. The input is the 13 channel inventory in every row, the rows and the splits and the ridge rule are the ones of Appendix~\ref{app:protocol}, and the only thing that moves is the width of the net that reads the channels. The parameter count runs from 13 to $5.4 \cdot 10^{5}$.

The inflation rises and settles. The whole axis is worth $1.61$ AUROC, the step out of the linear fit takes $0.60$ of that, and the last order of magnitude takes $0.04$. The hold-out arm is the part that does not move: it sits at $90.15$ at 13 parameters and at $89.81$ at $5.4 \cdot 10^{5}$, so the entire gain that the parameters buy stays inside the dataset the fit was given. At a fixed 13 parameters, exchanging what the fit reads moves the inflation by more than the whole parameter axis moves it, and the material block of Table~\ref{tab:capacity} carries that comparison. The first and the eighth row reproduce the 13 channel and the 13 channel MLP rows of Table~\ref{tab:capacity} exactly.

\begin{table}[t]
\centering
\caption{The parameter count as an axis, on the 13 channel inventory, over the 11 ImageNet backbones. Each row fits the same rows under the same splits and differs only in the width of the net. AUROC in percent, and the last column gives the standard deviation of the inflation across backbones.}
\label{tab:mlpwidth}
\small
\begin{tabular}{llrrrr}
\toprule
Fit & parameters & within & held out & inflation & sd \\
\midrule
ridge                  & $13$                       & $92.09$ & $90.15$ & $+1.94$ & $0.81$ \\
net, hidden $8$        & $153$                      & $92.62$ & $90.08$ & $+2.54$ & $0.94$ \\
net, hidden $16$       & $369$                      & $92.82$ & $90.03$ & $+2.79$ & $0.85$ \\
net, hidden $32$       & $993$                      & $93.02$ & $90.06$ & $+2.96$ & $0.82$ \\
net, hidden $64$       & $3009$                     & $93.18$ & $90.05$ & $+3.13$ & $0.82$ \\
net, hidden $128$      & $1.0\cdot10^{4}$           & $93.29$ & $89.93$ & $+3.36$ & $0.79$ \\
net, hidden $256$      & $3.7\cdot10^{4}$           & $93.36$ & $89.85$ & $+3.51$ & $0.76$ \\
net, hidden $512$      & $1.4\cdot10^{5}$           & $93.37$ & $89.84$ & $+3.53$ & $0.77$ \\
net, hidden $1024$     & $5.4\cdot10^{5}$           & $93.36$ & $89.81$ & $+3.55$ & $0.74$ \\
\bottomrule
\end{tabular}
\end{table}

\section{The variance of the class distances}
\label{app:identity}

The denominator of VarRatio has an exact expansion in the geometry of the class means. Write $\ell_c = z^{\top}\mu_c - \tfrac12 \lVert \mu_c \rVert^{2}$. Substituting $d_c = \lVert z \rVert^{2} - 2\ell_c$ into the variance removes the sample norm, because $\lVert z \rVert^{2}$ does not depend on $c$,
\begin{equation*}
\operatorname{Var}_c d_c \;=\; 4 \operatorname{Var}_c \ell_c .
\end{equation*}
Write $\bar{\mu}$ for the mean of the class means, $m_c = \lVert \mu_c \rVert^{2}$ and $\bar{m}$ for the mean of those norms. Expanding $\ell_c$ gives three terms,
\begin{align*}
\operatorname{Var}_c \ell_c &\;=\; \underbrace{z^{\top} C_{\mu} z}_{Q} \;-\; \underbrace{z^{\top} g}_{L} \;+\; \tfrac14 \underbrace{\operatorname{Var}_c m_c}_{T_3},\\[2pt]
C_{\mu} &\;=\; \operatorname*{mean}_c\, (\mu_c - \bar{\mu})(\mu_c - \bar{\mu})^{\top},
\qquad
g \;=\; \operatorname*{mean}_c\, (\mu_c - \bar{\mu})(m_c - \bar{m}),
\nonumber
\end{align*}
so that $\operatorname{Var}_c d_c = 4(Q - L) + T_3$. The third term depends only on the class means and takes the same value for every sample. The first two move with the sample. Their difference can go below zero, which happens on \ds{Textures} for several backbones. $T_3$ therefore acts as a reference level rather than as a bound.

The expansion gives a reading of what the denominator measures, namely how far the sample sits along the directions in which the class means themselves spread. A second decomposition makes that reading exact. Write $V$ for the span of $\{\mu_c - \bar{\mu}\}$ and split the centered sample as $z - \bar{\mu} = u + w$ with $u \in V$ and $w$ orthogonal to $V$. Every class mean is $\mu_c = \bar{\mu} + v_c$ with $v_c \in V$, so
\begin{equation*}
d_c \;=\; \lVert u - v_c \rVert^{2} \;+\; \lVert w \rVert^{2}
\qquad \text{for every } c,
\end{equation*}
and the second term takes the same value for every class. It cancels in $\operatorname{Var}_c d_c$ and survives in $d_{(1)}$. VarRatio therefore reads the position inside the class mean subspace against the distance in the full space, and the two terms of \eqref{eq:varratio} are not measuring the same geometry.

How much this bites depends on the benchmark. The subspace $V$ has dimension at most $\min(C - 1, D)$. On CIFAR-100 that is 99 against feature widths of 342 to 1280, so most of the whitened space leaves the denominator and stays in the numerator. On ImageNet-1k it is 999, and what stays outside $V$ depends on the backbone. Table~\ref{tab:residual} gives the residual width. It runs from nothing on the four backbones no wider than 768 to 1049 on the two ResNet-50 checkpoints, where more than half the whitened space never reaches the denominator. The score behaves the same way across that whole range and on both benchmarks, so whatever VarRatio measures does not depend on this cancellation.

\begin{table}[t]
\centering
\caption{What stays outside the class mean subspace on ImageNet-1k. The subspace $V$ has dimension at most $\min(C-1, D)$, which is 999 here, so the residual width is $D - \min(999, D)$.}
\label{tab:residual}
\small
\begin{tabular}{rlr}
\toprule
$D$ & Backbones & residual width \\
\midrule
$512$  & ResNet-18 & $0$ \\
$768$  & ViT-B/16, ViT-B/16 SWAG, DeiT3-B/16 & $0$ \\
$1024$ & Swin-B, ConvNeXt-B, ConvNeXtV2-B, SwinV2-B & $25$ \\
$1280$ & MobileNetV3-L & $281$ \\
$2048$ & ResNet-50 V1, ResNet-50 V2 & $1049$ \\
\bottomrule
\end{tabular}
\end{table}

\section{The exponent and the floor}
\label{app:exponent}

Section~\ref{sec:score} fixes the exponent of the denominator at 1. The general family with two constants reads
\begin{equation}
S_{\beta,\tau}(x) \;=\; \log d_{(1)}(x) \;-\; \beta \log\!\big(4(Q - L) + \tau\big),
\label{eq:betatau}
\end{equation}
where $\beta = 0$ recovers Mahalanobis++, $\beta = 1$ with $\tau = T_3$ recovers VarRatio and $\beta = \tfrac12$ is the only dimensionally consistent choice. We sweep both constants against the test sets themselves, which is the most generous condition available to them.

Table~\ref{tab:betatau} gives the sweep. It buys almost nothing. The joint optimum sits a median of $0.19$ AUROC above the derived point, and 6 of the 11 backbones land within $0.5$ of it. Any exponent leaves the score scale free, because $z \to \lambda z$ shifts $S_{\beta,\tau}$ by $(2 - 4\beta)\log\lambda$ for every sample at once, so the choice is not a question of units. The two constants also trade against each other along a ridge. Here $\beta_{\text{opt}}$ spreads over the whole grid while the score barely moves, and across backbones the two optima rise together at a rank correlation of $0.66$, so neither value carries meaning on its own. Publishing either one would reintroduce the reporting problem of Section~\ref{sec:prelim}.

\begin{table}[t]
\centering
\caption{Sweeping the two constants of \eqref{eq:betatau} against the test sets themselves, on the 11 ImageNet backbones. All columns are read from one grid over $\beta \in [0, 2]$ and $\tau / T_3 \in [10^{-2.5}, 10]$. The $\beta = 1$ column is the derived point $(\beta = 1, \tau = T_3)$ that Section~\ref{sec:score} fixes, $\beta = \tfrac12$ is the only dimensionally consistent member and $\beta = 0$ recovers Mahalanobis++. Gain subtracts the derived point from the joint optimum. AUROC in percent, averaged over the 5 OOD sets.}
\label{tab:betatau}
\small
\begin{tabular}{lrrrrrrr}
\toprule
 & $\beta = 0$ & $\beta = \tfrac12$ & $\beta = 1$ & joint & & & \\
Backbone & ($d_{(1)}$) & (dim.\ cons.) & (derived) & optimum & gain & $\beta_{\text{opt}}$ & $\tau_{\text{opt}}/T_3$ \\
\midrule
ConvNeXtV2-B   & $94.53$ & $94.56$ & $94.58$ & $94.60$ & $+0.03$ & $0.30$ & $0.079$ \\
SwinV2-B       & $92.36$ & $92.52$ & $92.64$ & $92.82$ & $+0.19$ & $1.80$ & $0.596$ \\
DeiT3-B/16     & $91.20$ & $91.50$ & $91.74$ & $92.28$ & $+0.54$ & $1.70$ & $0.398$ \\
ResNet-50 V1   & $84.94$ & $86.88$ & $88.18$ & $89.44$ & $+1.26$ & $1.10$ & $0.398$ \\
Swin-B         & $88.40$ & $88.64$ & $88.86$ & $89.97$ & $+1.11$ & $2.00$ & $0.266$ \\
ViT-B/16       & $88.39$ & $88.46$ & $88.51$ & $88.58$ & $+0.07$ & $0.30$ & $0.079$ \\
ResNet-50 V2   & $87.56$ & $87.66$ & $87.76$ & $88.51$ & $+0.75$ & $1.50$ & $0.119$ \\
ViT-B/16 SWAG  & $91.33$ & $91.57$ & $91.73$ & $91.87$ & $+0.13$ & $0.90$ & $0.398$ \\
ConvNeXt-B     & $88.19$ & $88.43$ & $88.66$ & $90.30$ & $+1.64$ & $1.70$ & $0.119$ \\
MobileNetV3-L  & $72.01$ & $79.51$ & $81.45$ & $81.59$ & $+0.14$ & $0.30$ & $0.119$ \\
ResNet-18      & $81.87$ & $84.75$ & $85.78$ & $85.91$ & $+0.12$ & $0.50$ & $0.119$ \\
\midrule
Mean           & $87.34$ & $88.59$ & $89.08$ & $89.63$ & $+0.54$ & $1.10$ & $0.119$ \\
\bottomrule
\end{tabular}
\end{table}

\section{Units of the mixing constant}
\label{app:units}

Section~\ref{sec:prelim} showed that $\alpha$ carries units of $L^{-2}$. Multiplying it by the ratio $\mathbb{E}_{\text{train}}[\operatorname{Var}_c d_c] / \mathbb{E}_{\text{train}}[d_{(1)}]$ removes those units and gives the dimensionless form
\begin{equation}
\alpha \;=\; \tilde{\alpha} \cdot \frac{\mathbb{E}_{\text{train}}\big[d_{(1)}\big]}{\mathbb{E}_{\text{train}}\big[\operatorname{Var}_c d_c\big]}.
\label{eq:dimensionless}
\end{equation}

Three measurements support the units argument. The first rescales the whitened space by hand. Sending $z \to \lambda z$ multiplies $d_{(1)}$ by $\lambda^{2}$ and $\operatorname{Var}_c d_c$ by $\lambda^{4}$, so the score $d_{(1)} - \alpha \operatorname{Var}_c d_c$ depends on $\alpha$ only through $\alpha \lambda^{2}$ and the optimum must move as $\lambda^{-2}$. It does, on 10 of 11 backbones at $\lambda = 0.5$ and on 10 of 11 at $\lambda = 2$. The median optimum lands at $0.06261$ against a prediction of $0.05257$ and at $0.00335$ against $0.00329$.

The second varies the covariance estimator instead of the coordinates, which rules out the one construction choice that could have produced the spread on its own. We stream the second moment once per checkpoint and then apply shrinkage strengths of $0$, $10^{-3}$ and $10^{-2}$. Table~\ref{tab:shrink} gives the result on five checkpoints. Shrinkage changes almost nothing. The score at $\alpha = 0$ moves by at most $0.14$ AUROC and the ratio moves by less than two percent. Within one backbone the dimensionless product $\tilde{\alpha}$ stays fixed to the resolution of the grid, which is what \eqref{eq:rescale} predicts. Across backbones it does not stay fixed, and that is the whole difficulty.

\begin{table}[t]
\centering
\caption{Shrinkage is not what moves the constant. AUROC on \ds{OpenImage-O}, which is the setting that the constant was published for. The ratio column reads the ID validation rows, while \eqref{eq:dimensionless} and Table~\ref{tab:dimless} read the training rows. The last column gives the dimensionless product at shrinkage $0$, $10^{-3}$ and $10^{-2}$.}
\label{tab:shrink}
\small
\begin{tabular}{lrrrl}
\toprule
Checkpoint & $\operatorname{Var}_c d_c / d_{(1)}$ & AUROC at $\alpha = 0$ & $\alpha^{*}$ & $\tilde{\alpha}$ \\
\midrule
ConvNeXtV2-B & $555.0 \to 544.6$ & $98.22 \to 98.21$ & $0.0041$ & 2.3 / 1.9 / 2.2 \\
SwinV2-B     & $360.8 \to 353.1$ & $97.95 \to 97.95$ & $0.0041$ & 1.5 / 1.8 / 1.7 \\
DeiT3-B/16   & $418.6 \to 410.9$ & $96.66 \to 96.66$ & $0.0073$ & 3.1 / 3.1 / 3.7 \\
ResNet-50 V1 & $24.2 \to 24.1$   & $90.40 \to 90.54$ & $0.0761$ & 1.8 / 1.8 / 1.8 \\
ResNet-50 V2 & $1650.7 \to 1568.0$ & $90.55 \to 90.64$ & $0.0073$ & 12.1 / 12.1 / 11.5 \\
\bottomrule
\end{tabular}
\end{table}

The third pair of checkpoints makes the practical point. ResNet-50 V1 and ResNet-50 V2 share an architecture and a feature width of 2048 and differ only in their training recipe. The ratio $\operatorname{Var}_c d_c / d_{(1)}$ moves from $24.2$ to $1650.7$ and the best $\alpha$ moves from $0.0761$ to $0.0073$. Across the 11 backbones of Table~\ref{tab:dimless} the ratio spans $6$ to $1623$. A single number published for one of them therefore does not carry to the other, while the dimensionless product lands between $1.5$ and $3.1$ for four of the five.

Rewriting the constant in the dimensionless form of \eqref{eq:dimensionless} repairs it. Table~\ref{tab:dimless} takes the value that the MahaVar authors recommend, multiplies it by the ratio of the checkpoint that they recommended it for and applies the result to all 11 ImageNet backbones. The rewritten constant beats Mahalanobis++ on 11 of 11 with a worst case of $+0.03$, and it lands $0.05$ below the zero constant reference on average. Choosing one shared dimensionless value on the test sets themselves gains a further $0.04$. Choosing it honestly, by holding out the backbone that it will be applied to, gives back $0.20$ and drops the worst case against the zero constant reference to $-2.36$. So a dimensionless constant transports where a dimensional one does not, and a constant that nobody has to publish still costs nothing to run.

\begin{table}[t]
\centering
\caption{The mixing constant made dimensionless, on ImageNet-1k. The ratio column gives the factor that turns $\alpha$ into $\tilde{\alpha}$ in \eqref{eq:dimensionless}. Published applies the value from the MahaVar paper after rewriting it. Shared picks one dimensionless value on the test sets and held out picks it from the other 10 backbones. AUROC in percent, averaged over the 5 OOD sets.}
\label{tab:dimless}
\small
\begin{tabular}{lrrrrrr}
\toprule
Backbone & ratio & Maha++ & VarRatio & published & shared & held out \\
\midrule
ConvNeXtV2-B     &   606.0 & 94.53 & 94.58 & 94.56 & 94.57 & 94.57 \\
SwinV2-B         &   375.4 & 92.36 & 92.64 & 92.65 & 92.75 & 92.70 \\
DeiT3-B/16       &   485.6 & 91.20 & 91.74 & 91.64 & 91.78 & 91.70 \\
ResNet-50 V1     &    24.0 & 84.94 & 88.18 & 88.28 & 88.88 & 88.60 \\
Swin-B           &  1603.4 & 88.40 & 88.86 & 88.74 & 88.87 & 88.80 \\
ViT-B/16         &   297.4 & 88.39 & 88.51 & 88.49 & 88.52 & 88.52 \\
ResNet-50 V2     &  1622.5 & 87.56 & 87.76 & 87.74 & 87.81 & 87.77 \\
ViT-B/16 SWAG    &   189.2 & 91.33 & 91.73 & 91.75 & 91.86 & 91.81 \\
ConvNeXt-B       &  1123.2 & 88.19 & 88.66 & 88.49 & 88.61 & 88.54 \\
MobileNetV3-L    &     6.2 & 72.01 & 81.45 & 81.25 & 80.54 & 79.09 \\
ResNet-18        &     9.5 & 81.87 & 85.78 & 85.71 & 85.55 & 85.55 \\
\midrule
Mean             &         & 87.34 & 89.08 & 89.03 & 89.07 & 88.88 \\
\bottomrule
\end{tabular}
\end{table}

\begin{figure}[htbp]
\centering
\includegraphics[width=\textwidth]{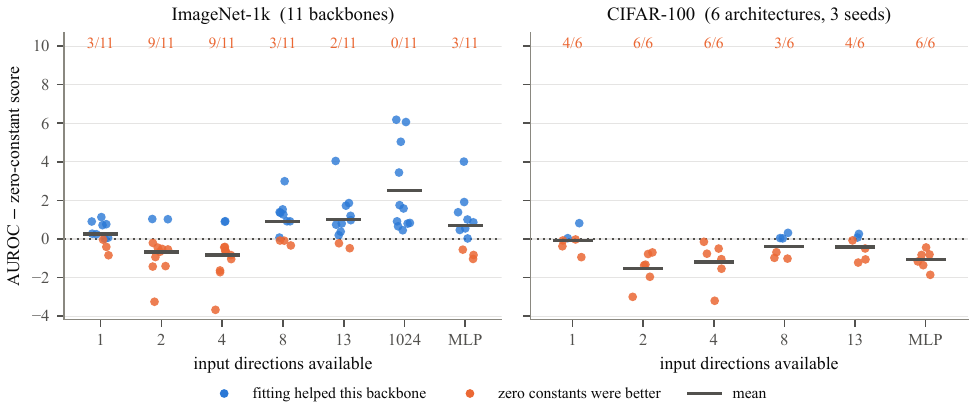}
\caption{Per backbone outcome of a fit that never sees the dataset it is scored on. Each dot subtracts the zero capacity reference from the AUROC of the held-out protocol for one backbone, and the bar gives the mean. The number above each column counts the backbones for which the zero capacity score came out ahead.}
\label{fig:loofail}
\end{figure}

\section{Full results}
\label{app:full}

Figure~\ref{fig:loofail} resolves the capacity curve by backbone. The tables below give every detector on every OOD dataset of every backbone, which is the resolution behind the averages of Table~\ref{tab:main}. Tables~\ref{tab:in-convnextv2-b-in21k} to \ref{tab:in-resnet18} cover ImageNet-1k and Tables~\ref{tab:c100-resnet18} to \ref{tab:c100-mobilenetv2} cover CIFAR-100. Every table reports AUROC and FPR@95 in percent. A star marks a detector whose constant we give an oracle value on the test sets. The CIFAR-100 tables average the 3 seeds of their architecture. VarRatio is the last row of every table. It fits nothing, so it is the reference that the other rows are read against.

\begin{table}[htbp]
\centering
\caption{ImageNet-1k with ConvNeXtV2-B (in21k).}
\label{tab:in-convnextv2-b-in21k}
\scriptsize\setlength{\tabcolsep}{2.5pt}
\begin{tabular}{lcccccccccccc}
\toprule
& \multicolumn{2}{c}{\ds{SSB-hard}} & \multicolumn{2}{c}{\ds{NINCO}} & \multicolumn{2}{c}{\ds{iNaturalist}} & \multicolumn{2}{c}{\ds{Textures}} & \multicolumn{2}{c}{\ds{OpenImage-O}} & \multicolumn{2}{c}{Avg} \\
Method & AUROC & FPR & AUROC & FPR & AUROC & FPR & AUROC & FPR & AUROC & FPR & AUROC & FPR \\
\midrule
MSP & 82.05 & 62.00 & 88.47 & 45.39 & 97.65 & 11.10 & 85.40 & 48.62 & 93.57 & 26.44 & 89.43 & 38.71 \\
MaxLogit & 84.08 & 52.96 & 88.76 & 34.91 & 98.26 & 6.14 & 85.85 & 41.01 & 94.50 & 18.02 & 90.29 & 30.61 \\
Energy & 84.54 & 49.24 & 88.60 & 33.00 & 98.32 & 5.56 & 85.71 & 39.47 & 94.73 & 15.68 & 90.38 & 28.59 \\
GEN$^{*}$ & 85.04 & 55.75 & 91.25 & 34.37 & 98.55 & 4.78 & 89.03 & 36.58 & 96.06 & 15.20 & 91.98 & 29.34 \\
KL-Matching & 78.96 & 66.86 & 88.28 & 49.42 & 96.57 & 15.20 & 87.40 & 48.49 & 93.56 & 29.98 & 88.96 & 41.99 \\
KNN$^{*}$ & 83.98 & 64.18 & 92.33 & 41.73 & 98.65 & 7.03 & 91.48 & 34.38 & 96.88 & 16.11 & 92.67 & 32.69 \\
ViM$^{*}$ & 87.98 & 49.20 & 94.01 & 31.54 & 99.73 & 1.10 & 92.07 & 31.13 & 98.20 & 9.05 & 94.40 & 24.41 \\
ReAct$^{*}$ & 85.83 & 49.05 & 91.61 & 32.10 & 98.99 & 4.62 & 87.87 & 38.28 & 96.58 & 13.88 & 92.18 & 27.59 \\
ORA & 80.69 & 62.84 & 90.83 & 41.80 & 98.83 & 5.78 & 89.87 & 36.76 & 96.88 & 16.19 & 91.42 & 32.67 \\
Mahalanobis++ & 87.11 & 49.17 & 95.09 & 25.35 & 99.68 & 1.16 & 92.55 & 29.22 & 98.21 & 9.14 & 94.53 & 22.81 \\
VarRatio & 87.22 & 48.64 & 95.23 & 24.57 & 99.70 & 1.10 & 92.51 & 29.31 & 98.22 & 9.15 & 94.58 & 22.55 \\
\bottomrule
\end{tabular}
\end{table}

\begin{table}[t]
\centering
\caption{ImageNet-1k with SwinV2-B (in21k).}
\label{tab:in-swinv2-b-in21k}
\scriptsize\setlength{\tabcolsep}{2.5pt}
\begin{tabular}{lcccccccccccc}
\toprule
& \multicolumn{2}{c}{\ds{SSB-hard}} & \multicolumn{2}{c}{\ds{NINCO}} & \multicolumn{2}{c}{\ds{iNaturalist}} & \multicolumn{2}{c}{\ds{Textures}} & \multicolumn{2}{c}{\ds{OpenImage-O}} & \multicolumn{2}{c}{Avg} \\
Method & AUROC & FPR & AUROC & FPR & AUROC & FPR & AUROC & FPR & AUROC & FPR & AUROC & FPR \\
\midrule
MSP & 79.62 & 65.72 & 86.64 & 49.95 & 94.76 & 21.94 & 84.69 & 50.30 & 91.63 & 33.00 & 87.47 & 44.18 \\
MaxLogit & 83.28 & 54.88 & 87.08 & 40.32 & 95.62 & 13.98 & 83.40 & 46.26 & 91.77 & 24.86 & 88.23 & 36.06 \\
Energy & 84.11 & 50.13 & 85.76 & 39.61 & 94.83 & 14.91 & 80.68 & 52.32 & 90.19 & 26.58 & 87.11 & 36.71 \\
GEN$^{*}$ & 82.11 & 61.18 & 90.11 & 40.49 & 97.24 & 10.67 & 88.93 & 37.66 & 95.13 & 19.31 & 90.70 & 33.86 \\
KL-Matching & 74.94 & 73.14 & 86.60 & 55.63 & 95.05 & 24.32 & 87.64 & 50.07 & 92.79 & 36.00 & 87.40 & 47.83 \\
KNN$^{*}$ & 74.47 & 82.31 & 88.22 & 56.62 & 98.26 & 7.46 & 90.27 & 38.51 & 95.91 & 21.37 & 89.43 & 41.25 \\
ViM$^{*}$ & 79.24 & 78.05 & 90.04 & 54.32 & 99.40 & 1.83 & 91.29 & 37.75 & 96.96 & 17.36 & 91.39 & 37.86 \\
ReAct$^{*}$ & 84.21 & 53.75 & 91.36 & 36.97 & 98.77 & 6.25 & 87.70 & 48.76 & 96.43 & 18.66 & 91.70 & 32.88 \\
ORA & 74.09 & 77.21 & 88.00 & 53.45 & 98.37 & 7.44 & 89.83 & 39.95 & 96.18 & 20.65 & 89.29 & 39.74 \\
Mahalanobis++ & 79.48 & 62.36 & 92.51 & 32.95 & 99.62 & 1.33 & 92.24 & 29.15 & 97.95 & 10.04 & 92.36 & 27.17 \\
VarRatio & 80.38 & 60.19 & 93.09 & 31.52 & 99.64 & 1.16 & 92.12 & 29.73 & 97.96 & 10.04 & 92.64 & 26.53 \\
\bottomrule
\end{tabular}
\end{table}

\begin{table}[t]
\centering
\caption{ImageNet-1k with DeiT3-B/16 (in21k).}
\label{tab:in-deit3-b16-in21k}
\scriptsize\setlength{\tabcolsep}{2.5pt}
\begin{tabular}{lcccccccccccc}
\toprule
& \multicolumn{2}{c}{\ds{SSB-hard}} & \multicolumn{2}{c}{\ds{NINCO}} & \multicolumn{2}{c}{\ds{iNaturalist}} & \multicolumn{2}{c}{\ds{Textures}} & \multicolumn{2}{c}{\ds{OpenImage-O}} & \multicolumn{2}{c}{Avg} \\
Method & AUROC & FPR & AUROC & FPR & AUROC & FPR & AUROC & FPR & AUROC & FPR & AUROC & FPR \\
\midrule
MSP & 74.18 & 75.74 & 80.65 & 64.15 & 90.20 & 38.50 & 75.50 & 67.55 & 87.20 & 48.21 & 81.54 & 58.83 \\
MaxLogit & 74.09 & 69.06 & 77.29 & 58.61 & 85.10 & 40.14 & 71.64 & 63.58 & 84.63 & 42.12 & 78.55 & 54.70 \\
Energy & 73.55 & 65.15 & 74.52 & 58.32 & 78.01 & 51.79 & 69.26 & 63.05 & 82.14 & 42.49 & 75.49 & 56.16 \\
GEN$^{*}$ & 77.95 & 67.56 & 85.99 & 50.99 & 95.36 & 18.09 & 81.66 & 53.28 & 93.00 & 27.46 & 86.79 & 43.48 \\
KL-Matching & 71.99 & 78.67 & 83.21 & 64.53 & 92.96 & 35.52 & 82.89 & 63.17 & 89.85 & 49.03 & 84.18 & 58.18 \\
KNN$^{*}$ & 76.64 & 78.95 & 88.61 & 56.48 & 97.37 & 11.85 & 88.78 & 47.34 & 95.42 & 25.09 & 89.36 & 43.94 \\
ViM$^{*}$ & 77.77 & 77.97 & 87.61 & 58.88 & 98.01 & 6.98 & 87.62 & 49.40 & 95.38 & 25.42 & 89.28 & 43.73 \\
ReAct$^{*}$ & 75.65 & 68.69 & 83.38 & 56.24 & 91.45 & 41.46 & 81.23 & 54.61 & 92.18 & 32.08 & 84.78 & 50.62 \\
ORA & 73.17 & 78.59 & 86.54 & 59.31 & 96.19 & 18.35 & 86.02 & 53.14 & 93.70 & 31.99 & 87.12 & 48.27 \\
Mahalanobis++ & 79.69 & 65.67 & 90.97 & 42.72 & 99.06 & 3.68 & 89.61 & 45.64 & 96.66 & 18.31 & 91.20 & 35.20 \\
VarRatio & 80.92 & 62.48 & 91.83 & 39.72 & 99.29 & 2.84 & 89.76 & 45.39 & 96.92 & 17.09 & 91.74 & 33.50 \\
\bottomrule
\end{tabular}
\end{table}

\begin{table}[t]
\centering
\caption{ImageNet-1k with ResNet-50 V1.}
\label{tab:in-resnet50}
\scriptsize\setlength{\tabcolsep}{2.5pt}
\begin{tabular}{lcccccccccccc}
\toprule
& \multicolumn{2}{c}{\ds{SSB-hard}} & \multicolumn{2}{c}{\ds{NINCO}} & \multicolumn{2}{c}{\ds{iNaturalist}} & \multicolumn{2}{c}{\ds{Textures}} & \multicolumn{2}{c}{\ds{OpenImage-O}} & \multicolumn{2}{c}{Avg} \\
Method & AUROC & FPR & AUROC & FPR & AUROC & FPR & AUROC & FPR & AUROC & FPR & AUROC & FPR \\
\midrule
MSP & 72.08 & 85.04 & 79.94 & 76.35 & 88.41 & 52.85 & 80.43 & 66.26 & 84.85 & 64.10 & 81.14 & 68.92 \\
MaxLogit & 72.49 & 84.35 & 80.39 & 76.62 & 91.16 & 50.72 & 86.38 & 54.89 & 89.13 & 57.89 & 83.91 & 64.90 \\
Energy & 72.07 & 84.45 & 79.69 & 77.66 & 90.63 & 53.72 & 86.72 & 52.50 & 89.02 & 57.73 & 83.62 & 65.21 \\
GEN$^{*}$ & 72.48 & 86.15 & 82.17 & 76.40 & 92.44 & 45.66 & 85.24 & 60.37 & 89.19 & 60.95 & 84.30 & 65.91 \\
KL-Matching & 69.63 & 82.68 & 80.70 & 73.31 & 89.90 & 46.09 & 82.68 & 65.04 & 86.31 & 61.22 & 81.84 & 65.67 \\
KNN$^{*}$ & 62.99 & 90.28 & 79.97 & 75.55 & 86.05 & 64.29 & 96.20 & 15.67 & 86.89 & 56.73 & 82.42 & 60.50 \\
ViM$^{*}$ & 65.54 & 90.21 & 78.59 & 77.56 & 89.61 & 59.19 & 97.53 & 11.06 & 90.52 & 50.84 & 84.36 & 57.77 \\
ReAct$^{*}$ & 73.01 & 80.74 & 81.72 & 70.09 & 96.34 & 19.82 & 91.62 & 41.49 & 91.83 & 41.57 & 86.90 & 50.74 \\
ORA & 71.76 & 85.03 & 84.14 & 70.16 & 94.45 & 34.76 & 94.25 & 27.38 & 92.14 & 45.93 & 87.35 & 52.65 \\
Mahalanobis++ & 65.08 & 89.78 & 79.45 & 75.08 & 90.86 & 49.44 & 98.79 & 5.67 & 90.54 & 46.30 & 84.94 & 53.25 \\
VarRatio & 70.25 & 84.97 & 83.58 & 68.15 & 95.71 & 25.02 & 98.89 & 5.07 & 92.46 & 37.27 & 88.18 & 44.10 \\
\bottomrule
\end{tabular}
\end{table}

\begin{table}[t]
\centering
\caption{ImageNet-1k with Swin-B.}
\label{tab:in-swin-b}
\scriptsize\setlength{\tabcolsep}{2.5pt}
\begin{tabular}{lcccccccccccc}
\toprule
& \multicolumn{2}{c}{\ds{SSB-hard}} & \multicolumn{2}{c}{\ds{NINCO}} & \multicolumn{2}{c}{\ds{iNaturalist}} & \multicolumn{2}{c}{\ds{Textures}} & \multicolumn{2}{c}{\ds{OpenImage-O}} & \multicolumn{2}{c}{Avg} \\
Method & AUROC & FPR & AUROC & FPR & AUROC & FPR & AUROC & FPR & AUROC & FPR & AUROC & FPR \\
\midrule
MSP & 71.30 & 80.34 & 78.51 & 71.55 & 86.47 & 51.06 & 78.71 & 65.25 & 81.68 & 62.42 & 79.33 & 66.13 \\
MaxLogit & 65.04 & 81.34 & 71.77 & 72.52 & 78.88 & 55.36 & 73.91 & 62.06 & 70.65 & 66.20 & 72.05 & 67.50 \\
Energy & 58.97 & 86.24 & 64.05 & 79.69 & 67.73 & 73.54 & 69.33 & 66.60 & 60.14 & 76.63 & 64.05 & 76.54 \\
GEN$^{*}$ & 74.21 & 78.79 & 83.03 & 64.26 & 92.01 & 34.24 & 84.63 & 53.10 & 87.53 & 49.00 & 84.28 & 55.88 \\
KL-Matching & 72.31 & 79.89 & 81.89 & 69.92 & 90.56 & 46.08 & 83.55 & 62.45 & 88.07 & 56.28 & 83.28 & 62.92 \\
KNN$^{*}$ & 72.23 & 86.39 & 83.49 & 73.61 & 92.22 & 52.79 & 88.57 & 48.74 & 91.59 & 54.00 & 85.62 & 63.11 \\
ViM$^{*}$ & 74.42 & 82.21 & 84.45 & 69.67 & 94.60 & 34.98 & 86.77 & 55.32 & 92.75 & 41.94 & 86.60 & 56.82 \\
ReAct$^{*}$ & 68.75 & 85.01 & 80.24 & 75.16 & 88.57 & 59.26 & 84.06 & 58.69 & 85.75 & 62.23 & 81.48 & 68.07 \\
ORA & 71.22 & 84.65 & 82.07 & 70.69 & 92.25 & 46.11 & 87.77 & 53.85 & 91.63 & 47.28 & 84.99 & 60.52 \\
Mahalanobis++ & 77.25 & 69.80 & 87.45 & 56.82 & 96.52 & 16.48 & 87.03 & 59.10 & 93.76 & 34.41 & 88.40 & 47.32 \\
VarRatio & 77.92 & 67.92 & 88.06 & 54.19 & 97.01 & 13.69 & 87.24 & 58.44 & 94.04 & 32.91 & 88.86 & 45.43 \\
\bottomrule
\end{tabular}
\end{table}

\begin{table}[t]
\centering
\caption{ImageNet-1k with ViT-B/16.}
\label{tab:in-vit-b-16}
\scriptsize\setlength{\tabcolsep}{2.5pt}
\begin{tabular}{lcccccccccccc}
\toprule
& \multicolumn{2}{c}{\ds{SSB-hard}} & \multicolumn{2}{c}{\ds{NINCO}} & \multicolumn{2}{c}{\ds{iNaturalist}} & \multicolumn{2}{c}{\ds{Textures}} & \multicolumn{2}{c}{\ds{OpenImage-O}} & \multicolumn{2}{c}{Avg} \\
Method & AUROC & FPR & AUROC & FPR & AUROC & FPR & AUROC & FPR & AUROC & FPR & AUROC & FPR \\
\midrule
MSP & 68.93 & 84.78 & 78.11 & 73.19 & 88.19 & 51.42 & 82.97 & 60.51 & 84.82 & 59.84 & 80.60 & 65.95 \\
MaxLogit & 64.19 & 85.68 & 72.38 & 73.83 & 85.27 & 52.16 & 81.64 & 56.61 & 81.53 & 58.77 & 77.00 & 65.41 \\
Energy & 58.78 & 88.19 & 66.00 & 78.62 & 79.28 & 63.94 & 79.24 & 58.46 & 76.44 & 64.85 & 71.95 & 70.81 \\
GEN$^{*}$ & 70.33 & 85.77 & 81.71 & 71.57 & 92.58 & 40.25 & 87.26 & 52.27 & 89.17 & 52.46 & 84.21 & 60.47 \\
KL-Matching & 69.17 & 83.71 & 81.59 & 71.98 & 90.35 & 50.18 & 85.31 & 63.00 & 87.67 & 59.74 & 82.82 & 65.72 \\
KNN$^{*}$ & 66.74 & 89.21 & 82.73 & 76.54 & 91.46 & 54.76 & 89.34 & 51.26 & 89.79 & 60.89 & 84.01 & 66.53 \\
ViM$^{*}$ & 70.99 & 73.01 & 85.75 & 54.32 & 97.16 & 12.93 & 88.60 & 47.71 & 92.40 & 38.29 & 86.98 & 45.25 \\
ReAct$^{*}$ & 63.01 & 88.26 & 75.41 & 78.68 & 85.99 & 65.11 & 84.63 & 57.13 & 84.22 & 64.62 & 78.65 & 70.76 \\
ORA & 66.41 & 88.24 & 81.48 & 73.85 & 90.75 & 53.82 & 88.40 & 54.63 & 89.57 & 57.29 & 83.32 & 65.56 \\
Mahalanobis++ & 74.08 & 69.60 & 88.56 & 52.28 & 97.19 & 11.59 & 89.04 & 55.62 & 93.07 & 40.07 & 88.39 & 45.83 \\
VarRatio & 74.22 & 69.41 & 88.72 & 51.96 & 97.32 & 11.28 & 89.05 & 56.01 & 93.24 & 39.44 & 88.51 & 45.62 \\
\bottomrule
\end{tabular}
\end{table}

\begin{table}[t]
\centering
\caption{ImageNet-1k with ResNet-50 V2.}
\label{tab:in-resnet50-v2}
\scriptsize\setlength{\tabcolsep}{2.5pt}
\begin{tabular}{lcccccccccccc}
\toprule
& \multicolumn{2}{c}{\ds{SSB-hard}} & \multicolumn{2}{c}{\ds{NINCO}} & \multicolumn{2}{c}{\ds{iNaturalist}} & \multicolumn{2}{c}{\ds{Textures}} & \multicolumn{2}{c}{\ds{OpenImage-O}} & \multicolumn{2}{c}{Avg} \\
Method & AUROC & FPR & AUROC & FPR & AUROC & FPR & AUROC & FPR & AUROC & FPR & AUROC & FPR \\
\midrule
MSP & 68.37 & 85.51 & 76.78 & 76.15 & 84.82 & 59.61 & 75.24 & 72.15 & 82.27 & 66.97 & 77.50 & 72.08 \\
MaxLogit & 65.60 & 86.52 & 72.91 & 78.68 & 80.39 & 67.26 & 69.56 & 75.55 & 77.71 & 71.05 & 73.23 & 75.81 \\
Energy & 54.16 & 95.16 & 54.74 & 94.57 & 51.19 & 98.09 & 48.57 & 95.62 & 53.92 & 95.79 & 52.52 & 95.85 \\
GEN$^{*}$ & 70.47 & 86.10 & 81.68 & 73.89 & 92.29 & 40.31 & 82.88 & 62.09 & 89.03 & 58.61 & 83.27 & 64.20 \\
KL-Matching & 69.49 & 84.38 & 81.89 & 73.32 & 91.11 & 45.67 & 86.08 & 56.79 & 88.95 & 57.04 & 83.50 & 63.44 \\
KNN$^{*}$ & 63.83 & 90.37 & 79.19 & 77.08 & 85.34 & 74.21 & 91.49 & 26.99 & 86.19 & 63.12 & 81.21 & 66.35 \\
ViM$^{*}$ & 60.00 & 88.74 & 73.59 & 79.38 & 93.35 & 30.97 & 93.11 & 22.59 & 89.87 & 44.31 & 81.99 & 53.20 \\
ReAct$^{*}$ & 52.62 & 95.90 & 53.53 & 95.51 & 53.81 & 97.56 & 54.45 & 93.62 & 57.00 & 95.05 & 54.28 & 95.53 \\
ORA & 58.11 & 91.87 & 71.03 & 86.36 & 84.30 & 72.93 & 84.97 & 61.86 & 84.28 & 71.61 & 76.54 & 76.93 \\
Mahalanobis++ & 75.28 & 70.03 & 86.55 & 60.92 & 91.98 & 48.43 & 93.35 & 26.61 & 90.64 & 53.24 & 87.56 & 51.85 \\
VarRatio & 75.46 & 69.36 & 86.78 & 59.19 & 92.45 & 45.03 & 93.31 & 26.52 & 90.81 & 51.88 & 87.76 & 50.40 \\
\bottomrule
\end{tabular}
\end{table}

\begin{table}[t]
\centering
\caption{ImageNet-1k with ViT-B/16 SWAG.}
\label{tab:in-vit-b16-swag}
\scriptsize\setlength{\tabcolsep}{2.5pt}
\begin{tabular}{lcccccccccccc}
\toprule
& \multicolumn{2}{c}{\ds{SSB-hard}} & \multicolumn{2}{c}{\ds{NINCO}} & \multicolumn{2}{c}{\ds{iNaturalist}} & \multicolumn{2}{c}{\ds{Textures}} & \multicolumn{2}{c}{\ds{OpenImage-O}} & \multicolumn{2}{c}{Avg} \\
Method & AUROC & FPR & AUROC & FPR & AUROC & FPR & AUROC & FPR & AUROC & FPR & AUROC & FPR \\
\midrule
MSP & 76.88 & 74.52 & 84.41 & 60.96 & 94.02 & 27.43 & 85.04 & 53.12 & 90.01 & 43.99 & 86.07 & 52.00 \\
MaxLogit & 77.57 & 69.70 & 86.44 & 51.21 & 97.21 & 13.16 & 86.23 & 44.84 & 93.33 & 28.09 & 88.16 & 41.40 \\
Energy & 77.03 & 71.24 & 86.10 & 52.16 & 97.36 & 12.62 & 85.84 & 46.65 & 93.40 & 27.98 & 87.95 & 42.13 \\
GEN$^{*}$ & 80.64 & 68.56 & 88.84 & 49.12 & 97.50 & 11.69 & 89.12 & 39.70 & 94.74 & 24.46 & 90.17 & 38.70 \\
KL-Matching & 74.65 & 73.77 & 84.78 & 58.29 & 94.25 & 27.55 & 87.13 & 49.63 & 90.96 & 40.87 & 86.35 & 50.02 \\
KNN$^{*}$ & 75.14 & 80.57 & 87.66 & 58.52 & 96.30 & 17.87 & 90.82 & 39.82 & 94.39 & 31.08 & 88.86 & 45.57 \\
ViM$^{*}$ & 76.48 & 73.33 & 90.24 & 43.67 & 99.25 & 3.23 & 89.43 & 41.15 & 96.13 & 19.73 & 90.31 & 36.22 \\
ReAct$^{*}$ & 77.02 & 71.12 & 86.11 & 52.21 & 97.36 & 12.53 & 85.83 & 46.63 & 93.41 & 27.92 & 87.95 & 42.08 \\
ORA & 72.41 & 78.39 & 85.78 & 56.94 & 95.91 & 19.49 & 88.31 & 44.36 & 93.40 & 31.55 & 87.16 & 46.15 \\
Mahalanobis++ & 78.26 & 66.40 & 91.57 & 39.83 & 98.91 & 4.75 & 91.50 & 38.32 & 96.41 & 20.72 & 91.33 & 34.00 \\
VarRatio & 79.68 & 63.12 & 92.33 & 37.27 & 99.10 & 3.71 & 91.01 & 42.29 & 96.55 & 19.61 & 91.73 & 33.20 \\
\bottomrule
\end{tabular}
\end{table}

\begin{table}[t]
\centering
\caption{ImageNet-1k with ConvNeXt-B.}
\label{tab:in-convnext-b}
\scriptsize\setlength{\tabcolsep}{2.5pt}
\begin{tabular}{lcccccccccccc}
\toprule
& \multicolumn{2}{c}{\ds{SSB-hard}} & \multicolumn{2}{c}{\ds{NINCO}} & \multicolumn{2}{c}{\ds{iNaturalist}} & \multicolumn{2}{c}{\ds{Textures}} & \multicolumn{2}{c}{\ds{OpenImage-O}} & \multicolumn{2}{c}{Avg} \\
Method & AUROC & FPR & AUROC & FPR & AUROC & FPR & AUROC & FPR & AUROC & FPR & AUROC & FPR \\
\midrule
MSP & 75.46 & 75.94 & 81.08 & 66.64 & 89.53 & 44.36 & 72.49 & 67.85 & 83.13 & 59.20 & 80.34 & 62.80 \\
MaxLogit & 72.92 & 76.69 & 76.43 & 69.67 & 82.61 & 57.67 & 63.28 & 73.24 & 73.39 & 69.03 & 73.73 & 69.26 \\
Energy & 65.56 & 84.28 & 64.03 & 82.77 & 55.71 & 90.42 & 49.03 & 88.56 & 52.69 & 89.49 & 57.40 & 87.10 \\
GEN$^{*}$ & 78.75 & 74.86 & 85.66 & 61.04 & 94.68 & 25.90 & 79.99 & 55.74 & 89.78 & 44.27 & 85.77 & 52.36 \\
KL-Matching & 74.17 & 80.09 & 83.87 & 68.73 & 92.29 & 43.76 & 86.37 & 58.53 & 90.11 & 51.86 & 85.36 & 60.59 \\
KNN$^{*}$ & 72.96 & 83.95 & 84.10 & 71.64 & 92.97 & 47.91 & 90.34 & 44.18 & 92.32 & 49.69 & 86.54 & 59.48 \\
ViM$^{*}$ & 71.54 & 82.23 & 84.41 & 67.63 & 94.98 & 32.54 & 91.63 & 38.90 & 93.45 & 39.54 & 87.20 & 52.17 \\
ReAct$^{*}$ & 73.67 & 84.43 & 81.20 & 75.45 & 87.65 & 68.61 & 74.66 & 73.16 & 82.80 & 73.60 & 79.99 & 75.05 \\
ORA & 73.01 & 82.52 & 84.44 & 67.01 & 94.23 & 34.23 & 89.06 & 47.71 & 92.55 & 43.44 & 86.66 & 54.98 \\
Mahalanobis++ & 76.88 & 69.97 & 87.29 & 58.68 & 94.93 & 27.47 & 88.63 & 56.83 & 93.20 & 40.32 & 88.19 & 50.65 \\
VarRatio & 77.45 & 68.43 & 87.88 & 56.41 & 95.60 & 22.44 & 88.79 & 55.69 & 93.59 & 37.13 & 88.66 & 48.02 \\
\bottomrule
\end{tabular}
\end{table}

\begin{table}[t]
\centering
\caption{ImageNet-1k with MobileNetV3-L.}
\label{tab:in-mobilenet-v3}
\scriptsize\setlength{\tabcolsep}{2.5pt}
\begin{tabular}{lcccccccccccc}
\toprule
& \multicolumn{2}{c}{\ds{SSB-hard}} & \multicolumn{2}{c}{\ds{NINCO}} & \multicolumn{2}{c}{\ds{iNaturalist}} & \multicolumn{2}{c}{\ds{Textures}} & \multicolumn{2}{c}{\ds{OpenImage-O}} & \multicolumn{2}{c}{Avg} \\
Method & AUROC & FPR & AUROC & FPR & AUROC & FPR & AUROC & FPR & AUROC & FPR & AUROC & FPR \\
\midrule
MSP & 70.61 & 86.52 & 78.81 & 79.74 & 86.96 & 59.09 & 82.10 & 64.52 & 83.30 & 69.56 & 80.36 & 71.89 \\
MaxLogit & 68.30 & 88.97 & 78.88 & 81.05 & 91.07 & 54.33 & 89.03 & 53.53 & 88.64 & 64.16 & 83.18 & 68.41 \\
Energy & 67.65 & 89.83 & 78.17 & 82.07 & 90.72 & 56.02 & 89.24 & 52.34 & 88.53 & 64.45 & 82.86 & 68.94 \\
GEN$^{*}$ & 70.43 & 86.83 & 80.77 & 79.57 & 91.68 & 50.53 & 88.45 & 51.63 & 88.42 & 64.18 & 83.95 & 66.55 \\
KL-Matching & 68.94 & 85.06 & 80.00 & 77.31 & 89.46 & 48.71 & 83.49 & 64.36 & 85.77 & 62.96 & 81.53 & 67.68 \\
KNN$^{*}$ & 57.22 & 94.39 & 68.21 & 89.71 & 62.85 & 94.82 & 84.56 & 60.00 & 73.13 & 82.36 & 69.19 & 84.26 \\
ViM$^{*}$ & 55.36 & 96.84 & 70.40 & 90.23 & 74.57 & 91.73 & 75.51 & 83.21 & 77.77 & 83.37 & 70.72 & 89.08 \\
ReAct$^{*}$ & 67.65 & 89.83 & 78.17 & 82.07 & 90.72 & 56.02 & 89.24 & 52.32 & 88.53 & 64.45 & 82.86 & 68.94 \\
ORA & 64.46 & 90.84 & 78.99 & 81.22 & 86.54 & 71.69 & 90.76 & 44.68 & 87.90 & 61.90 & 81.73 & 70.07 \\
Mahalanobis++ & 58.13 & 93.67 & 69.31 & 88.72 & 67.88 & 92.61 & 88.68 & 46.28 & 76.03 & 79.27 & 72.01 & 80.11 \\
VarRatio & 68.67 & 80.70 & 76.22 & 76.40 & 92.62 & 40.50 & 87.08 & 53.76 & 82.67 & 61.20 & 81.45 & 62.51 \\
\bottomrule
\end{tabular}
\end{table}

\begin{table}[t]
\centering
\caption{ImageNet-1k with ResNet-18.}
\label{tab:in-resnet18}
\scriptsize\setlength{\tabcolsep}{2.5pt}
\begin{tabular}{lcccccccccccc}
\toprule
& \multicolumn{2}{c}{\ds{SSB-hard}} & \multicolumn{2}{c}{\ds{NINCO}} & \multicolumn{2}{c}{\ds{iNaturalist}} & \multicolumn{2}{c}{\ds{Textures}} & \multicolumn{2}{c}{\ds{OpenImage-O}} & \multicolumn{2}{c}{Avg} \\
Method & AUROC & FPR & AUROC & FPR & AUROC & FPR & AUROC & FPR & AUROC & FPR & AUROC & FPR \\
\midrule
MSP & 67.38 & 89.49 & 77.50 & 81.32 & 87.08 & 58.24 & 78.42 & 71.44 & 81.46 & 73.16 & 78.37 & 74.73 \\
MaxLogit & 65.45 & 89.37 & 76.48 & 81.29 & 90.28 & 54.24 & 85.56 & 56.91 & 86.31 & 66.28 & 80.82 & 69.62 \\
Energy & 64.81 & 89.46 & 75.57 & 81.78 & 89.82 & 56.13 & 86.25 & 53.03 & 86.35 & 65.02 & 80.56 & 69.08 \\
GEN$^{*}$ & 65.83 & 90.62 & 78.18 & 82.95 & 90.64 & 54.59 & 83.11 & 68.23 & 85.52 & 72.54 & 80.66 & 73.79 \\
KL-Matching & 65.64 & 85.17 & 78.19 & 77.65 & 88.51 & 52.75 & 80.71 & 72.73 & 83.06 & 69.20 & 79.22 & 71.50 \\
KNN$^{*}$ & 58.90 & 91.48 & 73.44 & 82.49 & 76.88 & 79.57 & 95.14 & 19.93 & 80.88 & 66.31 & 77.05 & 67.96 \\
ViM$^{*}$ & 60.12 & 93.24 & 72.42 & 86.53 & 76.95 & 90.99 & 92.31 & 39.04 & 82.81 & 74.48 & 76.92 & 76.86 \\
ReAct$^{*}$ & 64.64 & 87.80 & 77.40 & 76.37 & 94.00 & 34.94 & 91.16 & 40.41 & 89.23 & 52.65 & 83.29 & 58.43 \\
ORA & 68.11 & 86.33 & 81.47 & 73.73 & 92.38 & 43.02 & 92.60 & 32.62 & 89.84 & 52.98 & 84.88 & 57.74 \\
Mahalanobis++ & 62.83 & 90.55 & 75.75 & 78.36 & 86.18 & 63.61 & 97.27 & 12.45 & 87.31 & 56.13 & 81.87 & 60.22 \\
VarRatio & 68.29 & 83.44 & 79.32 & 70.28 & 94.87 & 28.24 & 96.79 & 14.61 & 89.64 & 44.67 & 85.78 & 48.25 \\
\bottomrule
\end{tabular}
\end{table}

\begin{table}[t]
\centering
\caption{CIFAR-100 with ResNet-18.}
\label{tab:c100-resnet18}
\scriptsize\setlength{\tabcolsep}{1.0pt}
\begin{tabular}{lcccccccccccccc}
\toprule
& \multicolumn{2}{c}{\ds{CIFAR-10}} & \multicolumn{2}{c}{\ds{TIN}} & \multicolumn{2}{c}{\ds{MNIST}} & \multicolumn{2}{c}{\ds{SVHN}} & \multicolumn{2}{c}{\ds{Textures}} & \multicolumn{2}{c}{\ds{Places365}} & \multicolumn{2}{c}{Avg} \\
Method & AUROC & FPR & AUROC & FPR & AUROC & FPR & AUROC & FPR & AUROC & FPR & AUROC & FPR & AUROC & FPR \\
\midrule
MSP & 79.59 & 78.18 & 80.95 & 75.04 & 76.78 & 88.16 & 82.70 & 73.37 & 78.39 & 79.08 & 77.72 & 80.38 & 79.35 & 79.04 \\
MaxLogit & 80.04 & 77.98 & 81.38 & 74.67 & 78.39 & 88.01 & 84.65 & 71.36 & 79.29 & 78.43 & 77.81 & 80.50 & 80.26 & 78.49 \\
Energy & 79.95 & 78.76 & 81.28 & 75.44 & 78.56 & 88.02 & 85.19 & 69.46 & 79.36 & 78.34 & 77.57 & 81.44 & 80.32 & 78.58 \\
GEN$^{*}$ & 80.02 & 78.62 & 81.40 & 75.23 & 78.46 & 88.11 & 85.04 & 69.73 & 79.40 & 78.28 & 77.78 & 81.14 & 80.35 & 78.52 \\
KL-Matching & 78.96 & 78.70 & 80.86 & 73.98 & 77.08 & 85.44 & 83.07 & 69.71 & 78.79 & 75.09 & 77.72 & 78.22 & 79.41 & 76.86 \\
KNN$^{*}$ & 78.05 & 79.12 & 82.10 & 73.24 & 80.80 & 85.23 & 85.29 & 65.20 & 81.37 & 72.15 & 78.02 & 78.86 & 80.94 & 75.63 \\
ViM$^{*}$ & 69.95 & 88.85 & 75.07 & 85.75 & 77.74 & 89.00 & 81.50 & 76.32 & 82.43 & 66.80 & 72.66 & 87.00 & 76.56 & 82.29 \\
ReAct$^{*}$ & 79.07 & 78.86 & 81.17 & 74.76 & 76.34 & 88.32 & 87.12 & 66.83 & 81.95 & 76.36 & 78.68 & 80.67 & 80.72 & 77.63 \\
ORA & 78.39 & 79.55 & 80.88 & 75.08 & 78.22 & 84.44 & 82.86 & 71.80 & 81.01 & 74.61 & 78.05 & 79.80 & 79.90 & 77.55 \\
Mahalanobis++ & 75.24 & 83.93 & 79.21 & 80.11 & 78.06 & 88.92 & 87.45 & 55.16 & 85.19 & 59.98 & 77.65 & 79.08 & 80.47 & 74.53 \\
VarRatio & 76.02 & 83.20 & 80.48 & 76.82 & 77.87 & 92.97 & 88.05 & 55.99 & 86.33 & 56.38 & 80.89 & 69.28 & 81.61 & 72.44 \\
\bottomrule
\end{tabular}
\end{table}

\begin{table}[t]
\centering
\caption{CIFAR-100 with ResNet-34.}
\label{tab:c100-resnet34}
\scriptsize\setlength{\tabcolsep}{1.0pt}
\begin{tabular}{lcccccccccccccc}
\toprule
& \multicolumn{2}{c}{\ds{CIFAR-10}} & \multicolumn{2}{c}{\ds{TIN}} & \multicolumn{2}{c}{\ds{MNIST}} & \multicolumn{2}{c}{\ds{SVHN}} & \multicolumn{2}{c}{\ds{Textures}} & \multicolumn{2}{c}{\ds{Places365}} & \multicolumn{2}{c}{Avg} \\
Method & AUROC & FPR & AUROC & FPR & AUROC & FPR & AUROC & FPR & AUROC & FPR & AUROC & FPR & AUROC & FPR \\
\midrule
MSP & 78.41 & 79.56 & 79.73 & 77.31 & 71.24 & 90.41 & 78.56 & 82.53 & 78.10 & 80.67 & 77.50 & 79.99 & 77.26 & 81.74 \\
MaxLogit & 78.91 & 79.12 & 80.35 & 76.51 & 72.81 & 90.38 & 79.55 & 82.68 & 79.45 & 79.73 & 77.54 & 79.99 & 78.10 & 81.40 \\
Energy & 78.91 & 79.22 & 80.39 & 76.61 & 73.09 & 90.11 & 79.56 & 83.51 & 79.68 & 78.81 & 77.44 & 80.39 & 78.18 & 81.44 \\
GEN$^{*}$ & 79.06 & 79.31 & 80.53 & 76.53 & 72.81 & 90.60 & 79.72 & 83.06 & 79.62 & 79.20 & 77.67 & 80.17 & 78.24 & 81.48 \\
KL-Matching & 77.77 & 80.08 & 79.59 & 77.09 & 71.89 & 89.05 & 78.63 & 80.77 & 78.12 & 78.94 & 77.23 & 79.81 & 77.21 & 80.96 \\
KNN$^{*}$ & 78.68 & 79.73 & 82.01 & 74.83 & 74.66 & 89.73 & 82.61 & 75.61 & 80.77 & 76.55 & 78.71 & 78.78 & 79.57 & 79.21 \\
ViM$^{*}$ & 75.84 & 82.80 & 80.49 & 76.62 & 75.52 & 84.58 & 84.79 & 68.96 & 85.72 & 60.08 & 77.45 & 78.74 & 79.97 & 75.30 \\
ReAct$^{*}$ & 78.52 & 79.61 & 80.35 & 76.00 & 71.25 & 89.87 & 82.24 & 80.16 & 81.35 & 75.81 & 78.64 & 78.68 & 78.72 & 80.02 \\
ORA & 78.21 & 79.97 & 80.68 & 75.87 & 72.13 & 89.61 & 80.77 & 78.63 & 81.02 & 75.50 & 78.53 & 79.14 & 78.56 & 79.79 \\
Mahalanobis++ & 78.07 & 80.15 & 81.26 & 75.69 & 74.93 & 86.70 & 84.66 & 66.82 & 84.54 & 63.43 & 79.05 & 74.82 & 80.42 & 74.60 \\
VarRatio & 77.95 & 81.02 & 81.77 & 74.08 & 75.01 & 90.44 & 86.27 & 63.65 & 84.89 & 63.65 & 81.19 & 66.63 & 81.18 & 73.25 \\
\bottomrule
\end{tabular}
\end{table}

\begin{table}[t]
\centering
\caption{CIFAR-100 with WRN-28-10.}
\label{tab:c100-wrn28-10}
\scriptsize\setlength{\tabcolsep}{1.0pt}
\begin{tabular}{lcccccccccccccc}
\toprule
& \multicolumn{2}{c}{\ds{CIFAR-10}} & \multicolumn{2}{c}{\ds{TIN}} & \multicolumn{2}{c}{\ds{MNIST}} & \multicolumn{2}{c}{\ds{SVHN}} & \multicolumn{2}{c}{\ds{Textures}} & \multicolumn{2}{c}{\ds{Places365}} & \multicolumn{2}{c}{Avg} \\
Method & AUROC & FPR & AUROC & FPR & AUROC & FPR & AUROC & FPR & AUROC & FPR & AUROC & FPR & AUROC & FPR \\
\midrule
MSP & 81.72 & 74.96 & 82.33 & 73.36 & 72.25 & 91.72 & 78.67 & 79.22 & 77.70 & 82.21 & 78.82 & 80.39 & 78.58 & 80.31 \\
MaxLogit & 81.83 & 75.33 & 82.62 & 73.46 & 73.83 & 91.32 & 79.82 & 78.27 & 78.19 & 82.13 & 78.57 & 81.15 & 79.14 & 80.28 \\
Energy & 81.74 & 75.91 & 82.59 & 73.77 & 74.38 & 89.81 & 80.27 & 76.16 & 78.30 & 81.74 & 78.24 & 82.40 & 79.25 & 79.96 \\
GEN$^{*}$ & 81.76 & 75.88 & 82.60 & 73.76 & 74.33 & 89.87 & 80.26 & 76.14 & 78.30 & 81.71 & 78.26 & 82.36 & 79.25 & 79.95 \\
KL-Matching & 81.38 & 76.11 & 82.41 & 72.60 & 72.52 & 91.12 & 79.12 & 76.83 & 78.33 & 79.18 & 79.08 & 78.75 & 78.81 & 79.10 \\
KNN$^{*}$ & 81.16 & 76.38 & 83.48 & 71.53 & 73.99 & 89.87 & 80.97 & 73.65 & 81.78 & 75.07 & 79.57 & 79.07 & 80.16 & 77.60 \\
ViM$^{*}$ & 74.80 & 88.12 & 80.21 & 80.12 & 79.56 & 79.68 & 86.07 & 59.52 & 92.37 & 36.49 & 77.56 & 81.09 & 81.76 & 70.84 \\
ReAct$^{*}$ & 81.66 & 75.91 & 82.62 & 73.70 & 74.91 & 89.82 & 83.04 & 76.15 & 80.03 & 81.66 & 78.71 & 82.35 & 80.16 & 79.93 \\
ORA & 80.24 & 76.50 & 81.52 & 73.25 & 66.53 & 91.05 & 80.22 & 76.39 & 81.38 & 74.21 & 80.39 & 76.60 & 78.38 & 78.00 \\
Mahalanobis++ & 81.14 & 75.04 & 82.67 & 73.64 & 74.57 & 87.87 & 81.29 & 73.86 & 84.77 & 66.55 & 79.39 & 78.95 & 80.64 & 75.98 \\
VarRatio & 81.72 & 73.15 & 82.93 & 73.44 & 74.75 & 92.18 & 80.95 & 80.24 & 83.41 & 75.06 & 79.94 & 78.52 & 80.62 & 78.76 \\
\bottomrule
\end{tabular}
\end{table}

\begin{table}[t]
\centering
\caption{CIFAR-100 with DenseNet-BC.}
\label{tab:c100-densenet-bc}
\scriptsize\setlength{\tabcolsep}{1.0pt}
\begin{tabular}{lcccccccccccccc}
\toprule
& \multicolumn{2}{c}{\ds{CIFAR-10}} & \multicolumn{2}{c}{\ds{TIN}} & \multicolumn{2}{c}{\ds{MNIST}} & \multicolumn{2}{c}{\ds{SVHN}} & \multicolumn{2}{c}{\ds{Textures}} & \multicolumn{2}{c}{\ds{Places365}} & \multicolumn{2}{c}{Avg} \\
Method & AUROC & FPR & AUROC & FPR & AUROC & FPR & AUROC & FPR & AUROC & FPR & AUROC & FPR & AUROC & FPR \\
\midrule
MSP & 77.97 & 79.86 & 78.02 & 77.85 & 76.81 & 84.37 & 78.50 & 77.60 & 74.08 & 82.80 & 75.89 & 81.31 & 76.88 & 80.63 \\
MaxLogit & 78.75 & 79.25 & 79.67 & 75.66 & 77.20 & 83.03 & 81.93 & 73.41 & 75.80 & 80.71 & 76.25 & 81.41 & 78.27 & 78.91 \\
Energy & 78.49 & 80.14 & 79.52 & 76.28 & 76.99 & 83.50 & 82.02 & 72.97 & 75.73 & 80.55 & 75.93 & 82.45 & 78.11 & 79.31 \\
GEN$^{*}$ & 79.10 & 79.62 & 79.89 & 75.90 & 78.15 & 83.67 & 82.45 & 72.49 & 76.20 & 80.70 & 76.63 & 81.78 & 78.74 & 79.03 \\
KL-Matching & 77.33 & 80.53 & 78.02 & 77.56 & 79.71 & 77.86 & 79.84 & 72.68 & 75.72 & 77.35 & 75.90 & 80.70 & 77.75 & 77.78 \\
KNN$^{*}$ & 74.02 & 84.57 & 79.37 & 77.51 & 91.32 & 47.54 & 90.72 & 40.06 & 91.00 & 37.91 & 72.68 & 83.74 & 83.19 & 61.89 \\
ViM$^{*}$ & 70.50 & 91.87 & 75.20 & 84.95 & 83.81 & 79.77 & 90.47 & 46.26 & 92.75 & 32.04 & 72.32 & 86.47 & 80.84 & 70.23 \\
ReAct$^{*}$ & 77.50 & 80.39 & 79.30 & 76.42 & 75.96 & 83.67 & 84.76 & 72.51 & 78.00 & 78.97 & 75.83 & 82.47 & 78.56 & 79.07 \\
ORA & 77.86 & 80.84 & 79.71 & 76.19 & 79.22 & 82.30 & 85.19 & 67.52 & 86.58 & 58.20 & 77.05 & 81.26 & 80.94 & 74.38 \\
Mahalanobis++ & 72.41 & 88.90 & 76.08 & 83.63 & 91.15 & 45.84 & 91.43 & 40.13 & 93.30 & 30.48 & 74.83 & 83.46 & 83.20 & 62.07 \\
VarRatio & 70.45 & 88.35 & 77.86 & 77.89 & 86.75 & 67.52 & 91.71 & 45.97 & 93.39 & 32.04 & 81.17 & 70.25 & 83.56 & 63.67 \\
\bottomrule
\end{tabular}
\end{table}

\begin{table}[t]
\centering
\caption{CIFAR-100 with VGG16-BN.}
\label{tab:c100-vgg16-bn}
\scriptsize\setlength{\tabcolsep}{1.0pt}
\begin{tabular}{lcccccccccccccc}
\toprule
& \multicolumn{2}{c}{\ds{CIFAR-10}} & \multicolumn{2}{c}{\ds{TIN}} & \multicolumn{2}{c}{\ds{MNIST}} & \multicolumn{2}{c}{\ds{SVHN}} & \multicolumn{2}{c}{\ds{Textures}} & \multicolumn{2}{c}{\ds{Places365}} & \multicolumn{2}{c}{Avg} \\
Method & AUROC & FPR & AUROC & FPR & AUROC & FPR & AUROC & FPR & AUROC & FPR & AUROC & FPR & AUROC & FPR \\
\midrule
MSP & 74.16 & 83.62 & 75.72 & 81.09 & 72.05 & 88.00 & 72.47 & 86.95 & 75.05 & 81.81 & 73.53 & 83.58 & 73.83 & 84.18 \\
MaxLogit & 72.77 & 82.98 & 74.72 & 78.74 & 73.34 & 85.37 & 74.83 & 83.02 & 76.33 & 76.01 & 70.85 & 82.26 & 73.81 & 81.40 \\
Energy & 72.72 & 83.18 & 74.76 & 78.45 & 73.76 & 84.69 & 75.28 & 81.62 & 76.65 & 74.31 & 70.72 & 82.17 & 73.98 & 80.74 \\
GEN$^{*}$ & 74.16 & 82.38 & 76.12 & 78.05 & 73.98 & 85.96 & 75.31 & 82.33 & 76.75 & 75.93 & 72.96 & 81.48 & 74.88 & 81.02 \\
KL-Matching & 74.32 & 83.68 & 76.42 & 81.14 & 74.74 & 86.50 & 72.95 & 87.14 & 75.20 & 82.55 & 74.32 & 83.51 & 74.66 & 84.09 \\
KNN$^{*}$ & 76.71 & 81.27 & 79.98 & 77.91 & 75.49 & 90.74 & 75.10 & 85.98 & 78.49 & 80.27 & 76.89 & 81.52 & 77.11 & 82.95 \\
ViM$^{*}$ & 76.43 & 81.98 & 79.38 & 80.04 & 75.48 & 92.18 & 75.44 & 85.91 & 77.71 & 83.92 & 75.92 & 83.87 & 76.73 & 84.65 \\
ReAct$^{*}$ & 72.82 & 83.19 & 75.36 & 78.11 & 74.28 & 81.99 & 73.90 & 81.94 & 76.81 & 74.21 & 72.49 & 81.10 & 74.28 & 80.09 \\
ORA & 75.93 & 81.75 & 78.06 & 78.68 & 74.62 & 88.43 & 73.59 & 86.51 & 76.59 & 80.74 & 75.67 & 82.00 & 75.74 & 83.02 \\
Mahalanobis++ & 76.40 & 82.99 & 78.96 & 80.13 & 75.02 & 90.19 & 75.84 & 81.52 & 78.03 & 79.60 & 76.46 & 80.56 & 76.78 & 82.50 \\
VarRatio & 76.36 & 83.96 & 79.94 & 77.95 & 76.86 & 91.11 & 76.93 & 83.74 & 79.61 & 76.71 & 79.07 & 72.14 & 78.13 & 80.93 \\
\bottomrule
\end{tabular}
\end{table}

\begin{table}[t]
\centering
\caption{CIFAR-100 with MobileNetV2.}
\label{tab:c100-mobilenetv2}
\scriptsize\setlength{\tabcolsep}{1.0pt}
\begin{tabular}{lcccccccccccccc}
\toprule
& \multicolumn{2}{c}{\ds{CIFAR-10}} & \multicolumn{2}{c}{\ds{TIN}} & \multicolumn{2}{c}{\ds{MNIST}} & \multicolumn{2}{c}{\ds{SVHN}} & \multicolumn{2}{c}{\ds{Textures}} & \multicolumn{2}{c}{\ds{Places365}} & \multicolumn{2}{c}{Avg} \\
Method & AUROC & FPR & AUROC & FPR & AUROC & FPR & AUROC & FPR & AUROC & FPR & AUROC & FPR & AUROC & FPR \\
\midrule
MSP & 72.91 & 84.68 & 75.23 & 81.64 & 65.23 & 94.39 & 63.61 & 88.47 & 70.33 & 84.95 & 72.88 & 84.30 & 70.03 & 86.41 \\
MaxLogit & 73.05 & 83.97 & 79.34 & 76.48 & 69.63 & 94.87 & 72.90 & 87.05 & 77.64 & 79.93 & 77.45 & 80.38 & 75.00 & 83.78 \\
Energy & 72.65 & 84.56 & 79.30 & 76.46 & 69.80 & 95.19 & 73.44 & 87.57 & 77.96 & 79.47 & 77.50 & 80.40 & 75.11 & 83.94 \\
GEN$^{*}$ & 73.14 & 84.27 & 79.47 & 76.38 & 69.71 & 95.09 & 73.38 & 86.95 & 77.87 & 79.65 & 77.41 & 80.51 & 75.16 & 83.81 \\
KL-Matching & 71.77 & 85.24 & 74.96 & 80.81 & 70.58 & 83.89 & 68.79 & 77.14 & 72.15 & 76.40 & 71.96 & 82.39 & 71.70 & 80.98 \\
KNN$^{*}$ & 67.93 & 88.27 & 72.55 & 85.86 & 87.61 & 63.83 & 81.10 & 67.18 & 80.39 & 70.68 & 61.80 & 91.08 & 75.23 & 77.82 \\
ViM$^{*}$ & 71.41 & 86.01 & 74.98 & 85.42 & 89.08 & 51.13 & 88.83 & 47.15 & 86.64 & 57.00 & 68.68 & 89.09 & 79.94 & 69.30 \\
ReAct$^{*}$ & 68.18 & 87.17 & 77.14 & 77.63 & 73.82 & 94.32 & 85.52 & 71.00 & 86.03 & 63.87 & 76.24 & 80.87 & 77.82 & 79.14 \\
ORA & 72.65 & 85.03 & 76.31 & 80.96 & 77.26 & 84.68 & 74.03 & 83.10 & 82.34 & 72.32 & 73.13 & 82.80 & 75.95 & 81.48 \\
Mahalanobis++ & 61.73 & 91.44 & 61.43 & 91.56 & 93.24 & 37.53 & 88.56 & 54.54 & 85.49 & 52.74 & 54.64 & 93.98 & 74.18 & 70.30 \\
VarRatio & 62.84 & 90.91 & 70.19 & 85.23 & 90.18 & 56.18 & 91.04 & 45.82 & 89.72 & 42.85 & 72.22 & 84.13 & 79.37 & 67.52 \\
\bottomrule
\end{tabular}
\end{table}

\end{document}

%% file: math_commands.tex
\usepackage{amsmath,amsfonts,bm}

\def\eqref#1{equation~\ref{#1}}

\def\1{\bm{1}}

\DeclareMathAlphabet{\mathsfit}{\encodingdefault}{\sfdefault}{m}{sl}
\SetMathAlphabet{\mathsfit}{bold}{\encodingdefault}{\sfdefault}{bx}{n}

